\documentclass{article}

\usepackage{iclr2027_conference,times}

\usepackage{amsmath,amsfonts,bm}

\def\eqref#1{equation~\ref{#1}}

\def\1{\bm{1}}

\DeclareMathAlphabet{\mathsfit}{\encodingdefault}{\sfdefault}{m}{sl}
\SetMathAlphabet{\mathsfit}{bold}{\encodingdefault}{\sfdefault}{bx}{n}

\usepackage{natbib}
\usepackage{hyperref}
\usepackage{url}
\usepackage{graphicx}
\usepackage{booktabs}
\usepackage{amsmath}
\usepackage{amssymb}
\usepackage{mathtools}
\usepackage{multirow}
\usepackage{microtype}
\usepackage{xcolor}
\usepackage{threeparttable}
\usepackage{makecell}
\usepackage{colortbl}
\usepackage{adjustbox}
\usepackage{tikz}
\usepackage{float}
\usepackage{placeins}
\usepackage{algorithm}
\usepackage{algpseudocode}
\usepackage{enumitem}
\usepackage{multicol}
\usepackage{amsthm}
\usepackage{tcolorbox}
\usepackage{wrapfig}
\usepackage{subcaption}
\usepackage{caption}
\usepackage{array}
\usepackage{etoolbox}

\tcbuselibrary{breakable,skins}

\definecolor{OursBoxBorder}{HTML}{C98F82}
\definecolor{OursBG}{HTML}{FFF7F5}
\definecolor{MetaGray}{HTML}{666666}

\tcbset{
  pinkdashbox/.style={
    enhanced,
    colback=OursBG,
    frame hidden,
    borderline={0.5pt}{0pt}{OursBoxBorder,dashed},
    sharp corners,
    left=5pt,
    right=5pt,
    top=2.5pt,
    bottom=2.5pt,
    before skip=4pt,
    after skip=4pt,
    fontupper=\small
  }
}

\newtcolorbox{takeawaybox}{
  pinkdashbox,
  breakable=false
}

\newtcolorbox{rqbox}{
  pinkdashbox,
  breakable=false
}

\setlist[itemize]{noitemsep,leftmargin=*,topsep=0pt}

\hypersetup{
    colorlinks=true,
    linkcolor=blue,
    citecolor=blue,
    urlcolor=blue
}

\makeatletter
\patchcmd{\@maketitle}
  {\LARGE\sc}
  {\fontsize{12.5}{14.5}\selectfont\bfseries}
  {}
  {}
\makeatother

\title{
\makebox[\textwidth][c]{%
\parbox{0.98\textwidth}{%
\centering
ANTMAN: Adaptive Need Tracking for Multi-Agent Navigation in Large Information Spaces
}}
}

\author{
Jerry Wang$^{1}$
\quad
Haibo Jin$^{1}$
\quad
Xiaopeng Yuan$^{1}$
\quad
Peng Kuang$^{1}$
\quad
Haohan Wang$^{1}$
\\[3pt]
{\small\normalfont
$^{1}$University of Illinois Urbana-Champaign
}
}

\iclrfinalcopy

\begin{document}

\maketitle


\lhead{Preprint. Under review.}


\begin{abstract}
Information-seeking agents increasingly operate over information spaces that are too large to process exhaustively. Yet many multi-agent systems organize computation around static partitions of the available space, causing coordination to grow with how information is segmented rather than with what the query still requires. We introduce \textsc{ANTMAN}, an adaptive coordination framework that treats evolving unresolved information needs as the unit of runtime coordination. \textsc{ANTMAN} maintains a revisable Need Graph that tracks unresolved requirements, accumulated evidence, prior attempts, and search progress, and uses this state to control worker selection, routing, and task-local recovery as new evidence is discovered. By separating the coordination policy from substrate-specific search interfaces, the same need-conditioned mechanism can operate across different information spaces. Experiments across multi-document question answering, controlled long-context scaling, and realistic structured navigation show that \textsc{ANTMAN} remains effective across settings, including when execution is delegated to substantially smaller worker models. Under a $16\times$ increase in searchable context, \textsc{ANTMAN} increases active coordination by only $1.23\times$, compared with more than $15\times$ for partition-driven baselines, while preserving strong answer quality.
\end{abstract}


\section{Introduction}
Modern information-seeking agents increasingly operate over large,
heterogeneous information spaces, gathering and reasoning over evidence
across long interaction trajectories
\citep{xi-etal-2026-survey,yao-etal-2026-arc,lee-etal-2026-prints}.
Yet access to more information does not necessarily translate into more
effective use of it. Performance can degrade as context grows even when
relevant evidence is successfully identified, and can remain sensitive to
where that evidence appears within the context
\citep{du-etal-2025-context,liu-etal-2024-lost}. This raises a natural question: \textbf{How can agents search increasingly large information spaces without
letting coordination grow with the space itself?}

\begin{figure}
    \centering
    \includegraphics[width=0.93\linewidth]{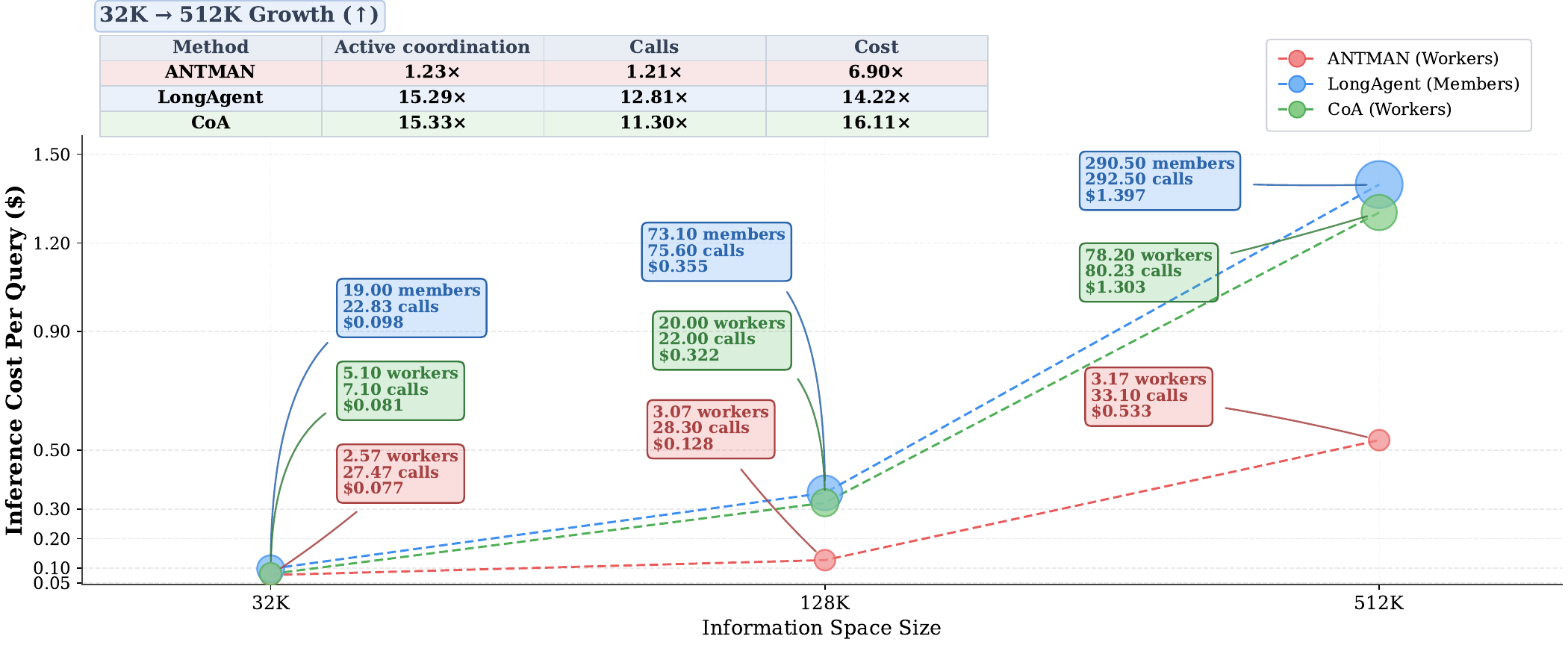}
    \caption{
    \textbf{ANTMAN decouples coordination from information-space growth.}
    Line height shows inference cost, bubble area shows active coordination, and annotations report model calls.
    ANTMAN maintains nearly constant coordination as the information space grows.
    }
    \label{fig:scaling}
\end{figure}

Distributing information processing across multiple agents can reduce the
context burden on any individual worker. Existing long-context multi-agent systems often follow this strategy by assigning different input partitions to different workers \citep{zhang2024chainagentslargelanguage}. For example, \textsc{LongAgent} partitions the input into fixed-size chunks and assigns each chunk to a separate member agent \citep{zhao-etal-2024-longagent}. This partition-based design reduces the context handled by any individual worker, but it also makes the size of the agent population grow with the number of partitions in the underlying information space.
This coupling exposes a mismatch between the organization of the information space and the amount of coordination a query actually requires. When the underlying information need remains fixed, expanding the available context should not by itself require a proportionally larger agent population. Yet static partition-based designs instantiate workers according to how the space is segmented, causing coordination to grow mechanically with context size rather than query demand.As Figure~\ref{fig:scaling} shows, under a $16\times$ increase in information-space size,
\textsc{ANTMAN}'s active coordination grows by only $1.23\times$, compared with
more than $15\times$ for partition-driven baselines, resulting in substantially
slower growth in model calls and inference cost. This motivates the question of what runtime state should determine the scope and
direction of coordination. We organize coordination around the information
requirements that remain unresolved as search progresses.

The relevant coordination state is itself dynamic. Classic work on information seeking has long argued that search is not simply a process of refining a fixed query; newly encountered information can change the searcher's understanding of the problem and redirect the search itself \citep{10.1108/eb024320}. For an information-seeking agent, new requirements may emerge only after intermediate evidence is discovered, previously identified needs may become resolved or require refinement, and unproductive search directions may need to be abandoned. Effective coordination therefore requires more than an initial decomposition of the query. It requires an explicit, revisable representation of what remains unresolved as evidence accumulates.

To address this challenge, we introduce \textsc{ANTMAN}, an adaptive coordination framework that treats evolving unresolved information needs, rather than input partitions, as the unit of runtime coordination. \textsc{ANTMAN} maintains a revisable \emph{Need Graph} as a runtime control state that tracks unresolved requirements, accumulated evidence, prior attempts, and search progress. The current Need Graph determines which needs should be pursued and which workers should become active; as new evidence arrives, the graph can resolve, refine, or introduce needs and redirect subsequent search. When progress stalls, \textsc{ANTMAN} invokes task-local recovery by reframing or rerouting unresolved needs, or by falling back to an alternative resolution path. By separating the coordination policy from substrate-specific search interfaces, the same need-conditioned coordination mechanism can operate across different information spaces.

We evaluate \textsc{ANTMAN} across three settings that test complementary consequences of this design: (1) multi-document question answering, which tests whether need-conditioned coordination preserves answer quality on standard information-seeking tasks; (2) controlled information-space scaling, which tests whether active coordination remains decoupled from irrelevant growth in the searchable space; and (3) realistic structured navigation, which tests whether the same coordination abstraction transfers beyond flat long-context inputs. Our main contributions are:
\begin{itemize}

     \item We introduce a revisable Need Graph as the runtime control state for
    multi-agent information seeking, allowing evolving unresolved requirements
    to govern worker activation, routing, and task-local recovery.
    Replacing this explicit state with graph-free adaptive replanning reduces
    performance by \textbf{23.82\%} at 512K and \textbf{16.67\%} on GAIA,
    demonstrating that the benefit extends beyond generic adaptive coordination.
\\
    \item We show that need-conditioned coordination transfers across distinct
information substrates without substrate-specific redesign.
ANTMAN outperforms the strongest non-benchmark-specific baselines by
\textbf{15.3\%} on RepoProbe, \textbf{18.4\%} on SWE-QA-Pro, and
\textbf{27.9\%} on GAIA. Also, the same coordination mechanism remains effective when
execution is delegated to substantially smaller \textbf{8B worker models}.
With Qwen3-8B for all non-orchestrator components, ANTMAN-H retains
\textbf{89.2\%}, \textbf{93.5\%}, and \textbf{98.3\%} of full ANTMAN's
performance on RepoProbe, SWE-QA-Pro, and GAIA, respectively.
\\
    \item We show that selective coordination preserves strong answer quality
    while substantially reducing sensitivity to evidence position.
    Across Early, Middle, and Late placements, ANTMAN has a middle-position gap
    of only $+.015$, compared with $-.181$ for direct full-context inference.

\end{itemize}

\definecolor{GroupBG}{HTML}{F1F3F6}
\definecolor{OursBand}{HTML}{F6E3DE}
\definecolor{OursBG}{HTML}{FCF0EC}
\definecolor{VenueGray}{HTML}{777777}
\definecolor{YesBG}{HTML}{EDF5EF}
\definecolor{PartialBG}{HTML}{F7F3E8}

\providecommand{\venue}[1]{%
    {\scriptsize\color{VenueGray}#1}%
}

\providecommand{\cmark}{\checkmark}

\newcommand{\yescell}{%
    \cellcolor{YesBG}\cmark
}

\newcommand{\partialcell}{%
    \cellcolor{PartialBG}$\triangle$
}

\newcommand{\nocell}{--}


\begin{table*}[t]
    \centering
    \small
    \renewcommand{\arraystretch}{1.12}
    \setlength{\tabcolsep}{4.0pt}

    \caption{
        \textbf{Comparison of representative information-seeking and
        coordination approaches.}
        \cmark\ denotes an explicit capability, $\triangle$ a partial one,
        and ``--'' a non-central property.
    }
    \label{tab:background_comparison}

    \begin{adjustbox}{max width=\textwidth}

    \begin{tabular}{
        @{}
        l
        c
        c
        c
        c
        c
        c
        @{}
    }

        \toprule

        \textbf{Method}
        &
        \makecell[c]{
            \textbf{Large / Structured}\\
            \textbf{Information Space}
        }
        &
        \makecell[c]{
            \textbf{Adaptive / Iterative}\\
            \textbf{Information Access}
        }
        &
        \makecell[c]{
            \textbf{Multi-Agent}\\
            \textbf{Collaboration}
        }
        &
        \makecell[c]{
            \textbf{Explicit Evolving}\\
            \textbf{Need / Gap State}
        }
        &
        \makecell[c]{
            \textbf{Need-Driven}\\
            \textbf{Coordination Scope}
        }
        &
        \makecell[c]{
            \textbf{Space-Decoupled}\\
            \textbf{Coordination}
        }
        \\

        \midrule


        \rowcolor{GroupBG}
        \multicolumn{7}{c}{
            \itshape Adaptive Information Seeking
        }
        \\

        Dense Retrieval (DPR)$^{\dagger}$
        ~\citep{karpukhin-etal-2020-dense}
        \;\venue{EMNLP'20}
        &
        \yescell
        &
        \nocell
        &
        \nocell
        &
        \nocell
        &
        \nocell
        &
        \nocell
        \\

        ChainRAG$^{\dagger}$
        ~\citep{zhu-etal-2025-mitigating}
        \;\venue{ACL'25}
        &
        \nocell
        &
        \yescell
        &
        \nocell
        &
        \partialcell
        &
        \partialcell
        &
        \nocell
        \\

        ReAct$^{\dagger}$
        ~\citep{yao2023reactsynergizingreasoningacting}
        \;\venue{ICLR'23}
        &
        \partialcell
        &
        \yescell
        &
        \nocell
        &
        \nocell
        &
        \nocell
        &
        \nocell
        \\

        ReAct + RepoGraph$^{\dagger}$
        ~\citep{ICLR2025_4a4a3c19}
        \;\venue{ICLR'25}
        &
        \yescell
        &
        \yescell
        &
        \nocell
        &
        \nocell
        &
        \nocell
        &
        \nocell
        \\

        RepoDistill$^{\dagger}$
        ~\citep{yin-etal-2026-repodistill}
        \;\venue{Findings ACL'26}
        &
        \yescell
        &
        \partialcell
        &
        \nocell
        &
        \nocell
        &
        \nocell
        &
        \nocell
        \\

        DRAGIN
        ~\citep{su-etal-2024-dragin}
        \;\venue{ACL'24}
        &
        \nocell
        &
        \yescell
        &
        \nocell
        &
        \partialcell
        &
        \partialcell
        &
        \nocell
        \\

        KiRAG
        ~\citep{fang-etal-2025-kirag}
        \;\venue{ACL'25}
        &
        \nocell
        &
        \yescell
        &
        \nocell
        &
        \partialcell
        &
        \partialcell
        &
        \nocell
        \\

        SelfRACG
        ~\citep{dong-etal-2025-selfracg}
        \;\venue{EMNLP'25}
        &
        \nocell
        &
        \yescell
        &
        \nocell
        &
        \partialcell
        &
        \partialcell
        &
        \nocell
        \\

        S2G-RAG
        ~\citep{li-etal-2026-s2g}
        \;\venue{ACL'26}
        &
        \nocell
        &
        \yescell
        &
        \nocell
        &
        \yescell
        &
        \partialcell
        &
        \nocell
        \\


        \midrule

        \rowcolor{GroupBG}
        \multicolumn{7}{c}{
            \itshape Multi-Agent Navigation and Coordination
        }
        \\

        LongAgent$^{\dagger}$
        ~\citep{zhao-etal-2024-longagent}
        \;\venue{EMNLP'24}
        &
        \yescell
        &
        \partialcell
        &
        \yescell
        &
        \nocell
        &
        \nocell
        &
        \nocell
        \\

        CoA$^{\dagger}$
        ~\citep{zhang2024chainagentslargelanguage}
        \;\venue{NeurIPS'24}
        &
        \yescell
        &
        \partialcell
        &
        \yescell
        &
        \nocell
        &
        \nocell
        &
        \nocell
        \\

        OWL$^{\dagger}$
        ~\citep{hu2025owl}
        \;\venue{NeurIPS'25}
        &
        \partialcell
        &
        \yescell
        &
        \yescell
        &
        \partialcell
        &
        \partialcell
        &
        \nocell
        \\

        C-3PO
        ~\citep{pmlr-v267-chen25an}
        \;\venue{ICML'25}
        &
        \nocell
        &
        \yescell
        &
        \yescell
        &
        \partialcell
        &
        \nocell
        &
        \nocell
        \\

        MAIN-RAG
        ~\citep{chang-etal-2025-main}
        \;\venue{ACL'25}
        &
        \nocell
        &
        \partialcell
        &
        \yescell
        &
        \nocell
        &
        \nocell
        &
        \nocell
        \\

        DyLAN
        ~\citep{liu2024a}
        \;\venue{COLM'24}
        &
        \nocell
        &
        \nocell
        &
        \yescell
        &
        \nocell
        &
        \nocell
        &
        \nocell
        \\


        \midrule

        \rowcolor{OursBand}
        \multicolumn{7}{c}{
            \itshape Ours
        }
        \\

        \rowcolor{OursBG}
        \textbf{ANTMAN}
        &
        \textbf{\cmark}
        &
        \textbf{\cmark}
        &
        \textbf{\cmark}
        &
        \textbf{\cmark}
        &
        \textbf{\cmark}
        &
        \textbf{\cmark}
        \\

        \bottomrule

    \end{tabular}

\end{adjustbox}

\vspace{2pt}

\begin{minipage}{0.99\textwidth}
    \scriptsize
    \color{VenueGray}
    $\dagger$ denotes methods used in our experiments. Need-driven coordination means unresolved needs control the scope
    of active computation, not only the next retrieval step. Space-decoupled coordination refers to whether active worker participation is structurally tied to the number of available information-space partitions.
\end{minipage}

\end{table*}

\section{Background and Motivation}
\label{sec:background}

Information-seeking agents must decide not only how to reason over evidence,
but also what information to access and how much computation to devote to
finding it.
Prior work addresses different parts of this problem through iterative
retrieval, structured navigation, and multi-agent coordination.
We review these directions before motivating ANTMAN's use of unresolved
information needs as the runtime state for controlling search and
coordination.


\subsection{Adaptive Information Seeking}
\label{sec:bg_information_seeking}

Retrieval methods reduce a larger corpus to a small set of candidate
evidence.
DPR~\citep{karpukhin-etal-2020-dense} provides a standard dense retrieval
baseline, while more recent methods make information access iterative.
ChainRAG progressively retrieves and rewrites across reasoning steps
\citep{zhu-etal-2025-mitigating}, and ReAct interleaves reasoning with
environment actions~\citep{yao2023reactsynergizingreasoningacting}.
For repository-level settings, RepoGraph exposes structural relations among
code entities~\citep{ICLR2025_4a4a3c19}, while RepoDistill combines
repository retrieval with learned context-budget allocation and compression
\citep{yin-etal-2026-repodistill}.
Recent systems further adapt retrieval according to what becomes necessary
during execution.
Several retrieval methods adapt information access according to evolving
information needs or gaps during execution
\citep{su-etal-2024-dragin,
       fang-etal-2025-kirag,
       dong-etal-2025-selfracg}.
Most directly, S2G-RAG judges whether accumulated evidence is sufficient
and, when it is not, generates structured gap items that become the next
retrieval query~\citep{li-etal-2026-s2g}. These methods show that information access can adapt as evidence accumulates
and that explicit needs or gaps can guide what information should be retrieved
next. ANTMAN builds on this idea but uses evolving unresolved needs as a
runtime control state for coordination. The same state governs not only what
information should be sought next, but also which workers should become
active and how search effort should be allocated.

\subsection{Multi-Agent Navigation and Coordination}
\label{sec:bg_multiagent}

Multi-agent systems distribute information processing across workers.
LongAgent partitions long inputs among member agents
\citep{zhao-etal-2024-longagent}, while Chain of Agents (CoA) processes
segmented long contexts through a sequence of collaborating workers
\citep{zhang2024chainagentslargelanguage}.
In both cases, the coordination footprint is closely tied to the
partitioning of the available input.
Other systems instead organize agents around specialized functions.
C-3PO and MAIN-RAG use multiple agents for retrieval and
evidence processing
\citep{pmlr-v267-chen25an,chang-etal-2025-main}.
OWL's Workforce combines hierarchical planning with coordinated
specialized tool-using workers and failure-triggered replanning
\citep{hu2025owl}.
Multi-agent organization can also adapt during execution through changes
in team composition, communication, or routing based on task and runtime
signals
\citep{liu2024a,
       wang-etal-2025-agentdropout,
       wang-etal-2025-megaagent,
       xiao-etal-2026-routerhgc}. Dynamic multi-agent coordination is not itself the contribution
of ANTMAN. The distinction instead lies in what runtime state governs this
adaptation. Prior work emphasizes either adaptive information
seeking, which changes what information to access, or adaptive multi-agent
coordination, which changes how computation is allocated.
ANTMAN connects these two forms of adaptation through a revisable
representation of unresolved information needs. Changes in what remains
unknown can therefore alter information access, worker activation, routing,
and  recovery during execution.
Table~\ref{tab:background_comparison} summarizes this distinction.

\definecolor{TheoryBlueBG}{HTML}{EEF5FC}
\definecolor{TheoryBlueBorder}{HTML}{AFCBE8}
\definecolor{TheoryBlueText}{HTML}{2F5F8F}

\newcounter{theorybox}

\section{Method}
\label{sec:method}

\begin{figure*}[t]
    \centering
    \includegraphics[width=\textwidth]{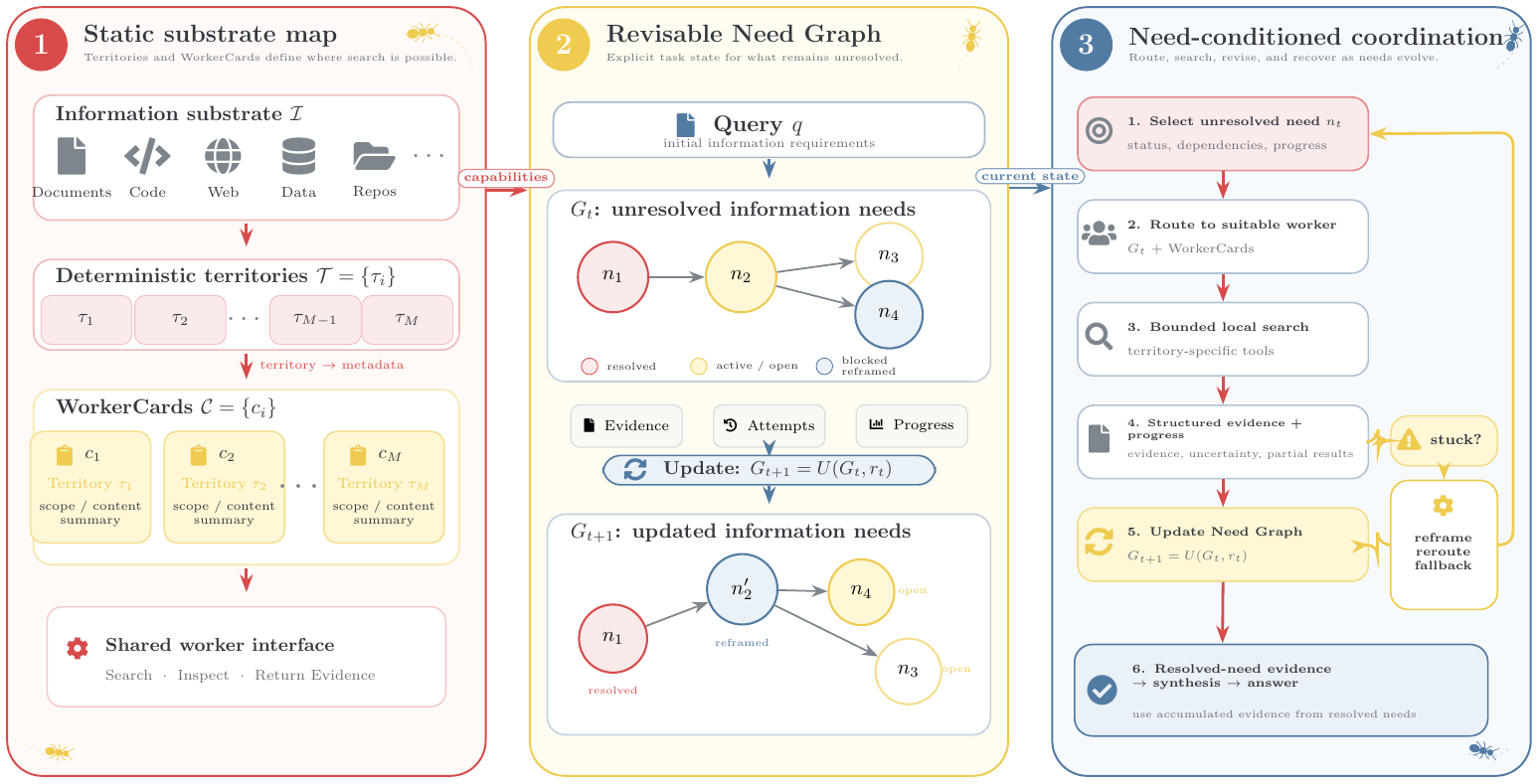}

    \caption{
        \textbf{Overview of ANTMAN.}
        \textbf{(1) Static substrate map:}
        the information substrate is partitioned into territories, each
        summarized by a WorkerCard describing its scope; workers operate
        through a shared substrate-specific interface.
        \textbf{(2) Revisable Need Graph:}
        unresolved information needs explicitly track evidence, prior
        attempts, and progress, and may be revised as new evidence arrives.
        \textbf{(3) Need-conditioned coordination:}
        the current Need Graph and WorkerCards determine worker routing;
        stalled needs can trigger reframing, rerouting, or fallback before
        final synthesis.
    }
    \label{fig:method}
\end{figure*}

ANTMAN maintains a revisable representation of unresolved information needs and coordinates workers as those needs evolve. Figure~\ref{fig:method} summarizes the workflow: \textbf{(1)} a static substrate map, \textbf{(2)} a runtime Need Graph, and \textbf{(3)} need-conditioned worker coordination.


\subsection{Problem Formulation}
\label{sec:problem_formulation}

We study information-seeking tasks in which answering a query $q$
requires locating and integrating evidence from a potentially large
information space $\mathcal{I}$.
The space may correspond to a document collection, long-context corpus,
software repository, web environment, or another tool-accessible
substrate. The central difficulty is that the amount of information available in
$\mathcal{I}$ can grow substantially while the information required by
$q$ remains comparatively small.
A coordination strategy that expands workers with the size of
$\mathcal{I}$ therefore couples computation to available information
rather than to the requirements of the query. ANTMAN instead makes evolving unresolved information needs the runtime control state for coordination, allowing worker activation and search decisions to adapt as those needs change.


\subsection{Static Substrate Map}
\label{sec:substrate_map}

Figure~\ref{fig:method}(1) describes \emph{where} ANTMAN can search.
The information space $\mathcal{I}$ is deterministically partitioned into
$M$ territories,
\begin{equation}
\mathcal{T}
=
P(\mathcal{I})
=
\{\tau_1,\ldots,\tau_M\},
\qquad
c_i=\mathrm{Card}(\tau_i),
\qquad
\mathcal{C}=\{c_1,\ldots,c_M\},
\label{eq:substrate_map}
\end{equation}
where $\tau_i$ is a bounded searchable region,
$P$ is the substrate-specific partitioning procedure,
$c_i$ is the WorkerCard associated with $\tau_i$, and
$\mathcal{C}$ is the complete WorkerCard set. A WorkerCard is lightweight territory metadata which summarizes the scope
and content represented by its assigned territory.
Worker specialization is therefore primarily \emph{scope-specialized}.
ANTMAN does not require different workers to use different model
architectures or tool sets.
Within a substrate, workers may share the same interface for searching,
inspecting, and returning evidence.

The icon strip in Figure~\ref{fig:method}(1) illustrates that the same
abstraction can be instantiated over different substrates; it does not
define a fixed set of worker personas.
Importantly, Equation~\ref{eq:substrate_map} defines the workers that are
available, not the workers that must participate in every query.


\subsection{Revisable Need Graph}
\label{sec:need_graph}

Figure~\ref{fig:method}(2) describes \emph{what} remains unresolved.
Given query $q$, ANTMAN initializes a Need Graph $G_0$ and maintains its
runtime state as
\begin{equation}
G_t=(N_t,\Delta_t),
\qquad
z_t(n)=
\bigl(
s_t(n),
\mathcal{E}_t(n),
\mathcal{H}_t(n),
p_t(n)
\bigr),
\label{eq:need_state}
\end{equation}
where $N_t$ is the set of information-need nodes at step $t$ and
$\Delta_t$ contains dependencies between them.
For a need $n\in N_t$, $s_t(n)$ denotes its resolution status,
$\mathcal{E}_t(n)$ its accumulated evidence,
$\mathcal{H}_t(n)$ its previous attempt history, and
$p_t(n)$ its current progress state.
These quantities correspond to the evidence, attempts, and progress
signals shown in Figure~\ref{fig:method}(2).

Unlike a fixed decomposition, $G_t$ is revised during execution. After a worker returns a structured report $r_t$, the coordinator updates the graph as $G_{t+1}=U(G_t,r_t)$, where $U$ is the task-local graph-update operator.
An update may resolve a need, preserve it as unresolved, reframe an
unsuccessful need into $n'$, or introduce additional dependencies or
requirements revealed by newly collected evidence.
The non-linear graph structure in Figure~\ref{fig:method}(2) reflects
that information requirements may branch or depend on one another rather
than forming a fixed sequential plan.


\subsection{Need-Conditioned Coordination and Recovery}
\label{sec:need_conditioned_coordination}

Figure~\ref{fig:method}(3) shows how the current graph controls
computation.
At coordination step $t$, ANTMAN selects an unresolved need $n_t$,
routes it using the current Need Graph and WorkerCards, and invokes the
corresponding territory-backed worker:
\begin{equation}
n_t=\mathrm{Select}(G_t),\qquad
i_t=\mathrm{Route}(n_t,G_t,\mathcal{C}),\qquad
r_t=\mathrm{Exec}(w_{i_t},n_t,\tau_{i_t}).
\label{eq:coordination_loop}
\end{equation}
where $i_t\in\{1,\ldots,M\}$ identifies the selected worker,
$w_{i_t}$ is the worker associated with territory $\tau_{i_t}$, and
$r_t$ is its structured report containing evidence, progress, and
remaining uncertainty.
The report is then fed back, closing the
runtime loop. Workers perform bounded local information seeking rather than
reconstructing the entire global task.
Global state remains in $G_t$, while a worker searches only within the
scope assigned to the selected need.
If repeated attempts fail to make progress, ANTMAN performs task-local
recovery by reframing the need, rerouting it to another territory, or
invoking a fallback resolution path.
Worker definitions and the static substrate map remain unchanged. Once the required needs are resolved, evidence attached to those needs is
used for final synthesis. The primary implementation prompts are provided in
Appendix~\ref{app:runtime_prompts}.


\paragraph{Need-conditioned coordination and space decoupling.}
Our design separates the number of \emph{available} territories from the
amount of \emph{active} coordination. Let $M$ denote the total number of
territories, $H(q)$ the number of information needs realized during execution,
and $\rho$ the maximum number of routing attempts per need. If $T_q$ denotes
the number of coordination steps for query $q$, then each step activates at
most one territory-backed worker, so
$|\mathcal{A}(q)| \leq T_q$. Since each realized need can be routed at most
$\rho$ times, $T_q \leq \rho H(q)$. Independently,
$|\mathcal{A}(q)| \leq M$, since no more than the $M$ available territories
can become active. These two constraints give the bound summarized in
Box~\ref{box:space_decoupling}. When realized information demand remains
bounded, enlarging the information space can increase $M$ without forcing
active coordination to grow with it.
Appendix~\ref{app:space_decoupling} formalizes this property, while
Section~\ref{sec:controlled_scaling} tests it by varying space size at
approximately fixed information demand.
\vspace{2pt}

\refstepcounter{theorybox}
\label{box:space_decoupling}
\begingroup
\setlength{\fboxsep}{4pt}
\setlength{\fboxrule}{0.5pt}
\noindent
\fcolorbox{TheoryBlueBorder}{TheoryBlueBG}{
    \parbox{0.96\linewidth}{
        \small
        \textcolor{TheoryBlueText}{
            \textbf{Box \thetheorybox: Space-Decoupling Bound}
        }
        \hfill
        $\displaystyle
        |\mathcal{A}(q)|
        \leq
        \min\!\left\{M,\rho H(q)\right\}.
        $
    }
}
\endgroup

\section{Experiments}

\definecolor{MiddleBG}{HTML}{F4F5F7}
\definecolor{OursMiddleBG}{HTML}{F8E8E4}

\definecolor{GroupBG}{HTML}{F1F3F6}
\definecolor{OursBand}{HTML}{F6E3DE}
\definecolor{OursBG}{HTML}{FCF0EC}
\definecolor{VenueGray}{HTML}{777777}
\definecolor{PendingGray}{HTML}{999999}
\definecolor{AblDropBG}{RGB}{255,237,235}
\definecolor{AblDropText}{RGB}{166,65,59}

\definecolor{AblGainBG}{RGB}{230,245,233}
\definecolor{AblGainText}{RGB}{39,113,68}

\newcommand{\AblDrop}[1]{%
    \colorbox{AblDropBG}{%
        \textcolor{AblDropText}{\scriptsize $\downarrow #1\%$}%
    }%
}

\newcommand{\AblGain}[1]{%
    \colorbox{AblGainBG}{%
        \textcolor{AblGainText}{\scriptsize $\uparrow #1\%$}%
    }%
}
\newcommand{\AblSame}[1]{\textcolor{gray}{(\(\leftrightarrow\) #1\%)}}
\newcommand{\pending}{%
    {\color{PendingGray}--}%
}

\subsection{Preliminary Evaluation on Standard Multi-Document QA}
\label{sec:preliminary_multidoc}

Before studying how ANTMAN behaves as information spaces grow, we first
evaluate its effectiveness in standard multi-document QA.
We consider HotpotQA~\citep{yang-etal-2018-hotpotqa},
2WikiMultiHopQA~\citep{ho-etal-2020-constructing}, and
MuSiQue~\citep{trivedi-etal-2022-musique}, which cover complementary
forms of cross-document and multi-hop reasoning.
We evaluate 30 questions per benchmark across nine methods, yielding
810 predictions scored with the same frozen clean-answer extraction and
evaluation pipeline.

\textbf{Model setting.}
All baselines and ANTMAN use GPT-4.1 \citep{openai2025gpt41}
throughout; ANTMAN-H retains the GPT-4.1 orchestrator but uses Qwen3-8B
\citep{qwen3technicalreport} for all non-orchestrator components.
\begin{table*}[t]
    \centering
    \scriptsize
    \renewcommand{\arraystretch}{0.92}
    \setlength{\tabcolsep}{2.8pt}

    \caption{
        \textbf{Performance on standard multi-document QA.}
        We report EM and token-level F1 over 30 questions per benchmark,
        with macro averages across the three benchmarks.
    }
    \label{tab:preliminary_multidoc}

    \begin{tabular*}{\textwidth}{
        l
        @{\extracolsep{\fill}}
        cc
        cc
        cc
        cc
    }

        \toprule

        \multirow{2}{*}{\textbf{Method}}
        &
        \multicolumn{2}{c}{
            \makecell[c]{
                \textbf{HotpotQA}\\[1.2pt]
            }
        }
        &
        \multicolumn{2}{c}{
            \makecell[c]{
                \textbf{2Wiki}\\[1.2pt]
            }
        }
        &
        \multicolumn{2}{c}{
            \makecell[c]{
                \textbf{MuSiQue}\\[1.2pt]
            }
        }
        &
        \multicolumn{2}{c}{\textbf{Avg.}}
        \\

        \cmidrule(lr){2-3}
        \cmidrule(lr){4-5}
        \cmidrule(lr){6-7}
        \cmidrule(lr){8-9}

        &
        EM & F1
        &
        EM & F1
        &
        EM & F1
        &
        EM & F1
        \\

        \midrule


        \rowcolor{GroupBG}
        \multicolumn{9}{c}{
            \itshape Retrieval and RAG Baselines
        }
        \\

        Iterative Sparse Retrieval (BM25)
        &
        .533 & .684
        &
        .367 & .515
        &
        .433 & .582
        &
        .444 & .594
        \\

        Dense Retrieval (DPR)
        ~\citep{karpukhin-etal-2020-dense}
        \;\venue{EMNLP'20}
        &
        .533 & .696
        &
        .267 & .379
        &
        .200 & .292
        &
        .333 & .456
        \\

        ChainRAG
        ~\citep{zhu-etal-2025-mitigating}
        \;\venue{ACL'25}
        &
        .567 & .711
        &
        .733 & .791
        &
        .500 & .654
        &
        .600 & .719
        \\

        S2G-RAG
        ~\citep{li-etal-2026-s2g}
        \;\venue{ACL'26}
        &
        \underline{.667} & \underline{.810}
        &
        \textbf{.800} & \textbf{.858}
        &
        \underline{.633} & \underline{.730}
        &
        \underline{.700} & \underline{.799}
        \\


        \rowcolor{GroupBG}
        \multicolumn{9}{c}{
            \itshape Agentic and Multi-Agent Baselines
        }
        \\

        ReAct
        ~\citep{yao2023reactsynergizingreasoningacting}
        \;\venue{ICLR'23}
        &
        .533 & .734
        &
        .567 & .781
        &
        .467 & .626
        &
        .522 & .714
        \\

        LongAgent
        ~\citep{zhao-etal-2024-longagent}
        \;\venue{EMNLP'24}
        &
        .433 & .655
        &
        .467 & .697
        &
        .400 & .494
        &
        .433 & .616
        \\

        CoA
        ~\citep{zhang2024chainagentslargelanguage}
        \;\venue{NeurIPS'24}
        &
        .567 & .769
        &
        .600 & .778
        &
        .533 & .682
        &
        .567 & .743
        \\


        \rowcolor{OursBand}
        \multicolumn{9}{c}{
            \itshape Ours
        }
        \\

        \rowcolor{OursBG}
        \textbf{ANTMAN-H$^\dagger$}
        &
        \textbf{.700} & \textbf{.842}
        &
        .700 & .796
        &
        .467 & .651
        &
        .622 & .763
        \\

        \rowcolor{OursBG}
        \textbf{ANTMAN}
        &
        \textbf{.700} & \textbf{.842}
        &
        \underline{.767} & \underline{.823}
        &
        \textbf{.667} & \textbf{.777}
        &
        \textbf{.711} & \textbf{.814}
        \\

        \bottomrule

    \end{tabular*}

    \vspace{1pt}

    \begin{minipage}{0.99\textwidth}
        \tiny
        \color{VenueGray}
        Best results are \textbf{bold}; second-best results are
        \underline{underlined}; ties share formatting.
        $^\dagger$ ANTMAN-H uses Qwen3-8B for all non-orchestrator
        components while retaining the same orchestrator as full ANTMAN.
    \end{minipage}

\end{table*}
\\
\textbf{Results.}
ANTMAN achieves the strongest aggregate performance despite being designed for large information spaces, remaining competitive with specialized retrieval systems such as S2G-RAG.
It performs best on HotpotQA and MuSiQue and remains competitive on 2WikiMultiHopQA.

\textbf{Smaller workers.}
ANTMAN-H remains competitive with all baselines except S2G-RAG despite using substantially smaller execution models.
It matches full ANTMAN on HotpotQA and remains close on 2WikiMultiHopQA, with a larger gap only on the more demanding MuSiQue benchmark.

\begin{takeawaybox}
\textbf{Preliminary takeaway.}
Need-driven coordination preserves strong multi-document QA
effectiveness rather than trading answer quality for scalability,
and remains effective even when execution is delegated to smaller
worker models.
\end{takeawaybox}
%
%
%
%
%


\subsection{Scaling and Navigation in Large Information Spaces}
\label{sec:large_information_spaces}

We study how ANTMAN behaves in large and structured information spaces,
asking whether need-driven coordination can remain efficient while still
locating the evidence required for each query. We examine this through two
research questions: \textbf{RQ1} tests whether active coordination remains
decoupled from information-space size when the underlying information need
is approximately fixed, while \textbf{RQ2} tests whether the same
coordination principle remains effective in realistic structured
environments that require navigation.


\subsubsection{Controlled Information-Space Scaling}
\label{sec:controlled_scaling}

\textbf{Setup.}
We follow the multi-needle setting of Needle-in-a-Haystack PLUS introduced by
LongAgent~\citep{zhao-etal-2024-longagent}. We evaluate 10 underlying questions at Early, Middle, and
Late evidence positions over 32K, 64K, 128K, and 512K contexts. For each
question, the required evidence is preserved while additional filler expands
the surrounding distractor space, yielding matched conditions that isolate
information-space growth while keeping the underlying information need
approximately fixed. We compare ANTMAN with LongAgent and CoA, two multi-agent systems whose
designs most closely match our setting by distributing a large information
space across multiple workers.


\begin{figure*}[t]
    \centering

    \begin{minipage}[t]{0.325\textwidth}
        \centering
        \includegraphics[
            width=\linewidth
        ]{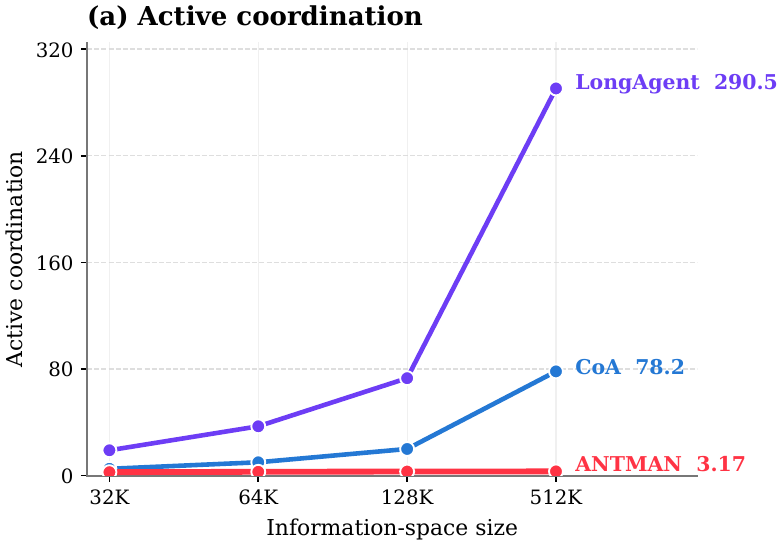}
    \end{minipage}
    \hfill
    \begin{minipage}[t]{0.325\textwidth}
        \centering
        \includegraphics[
            width=\linewidth
        ]{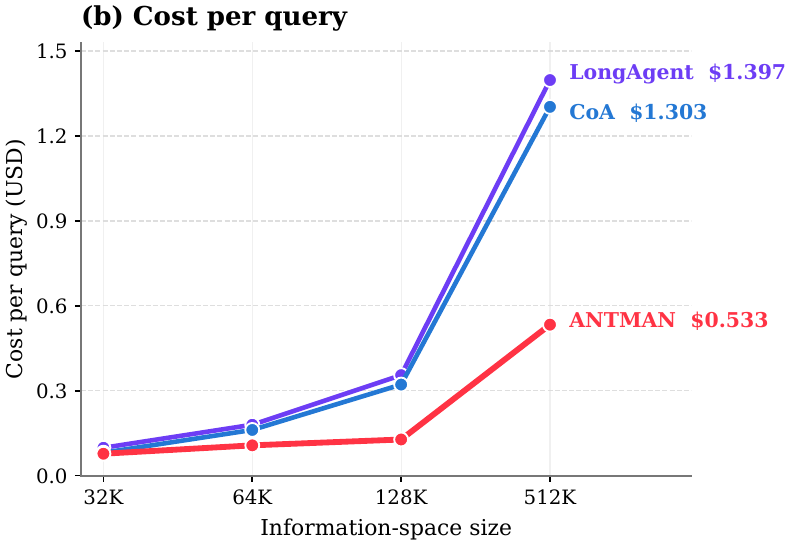}
    \end{minipage}
    \hfill
    \begin{minipage}[t]{0.325\textwidth}
        \centering
        \includegraphics[
            width=\linewidth
        ]{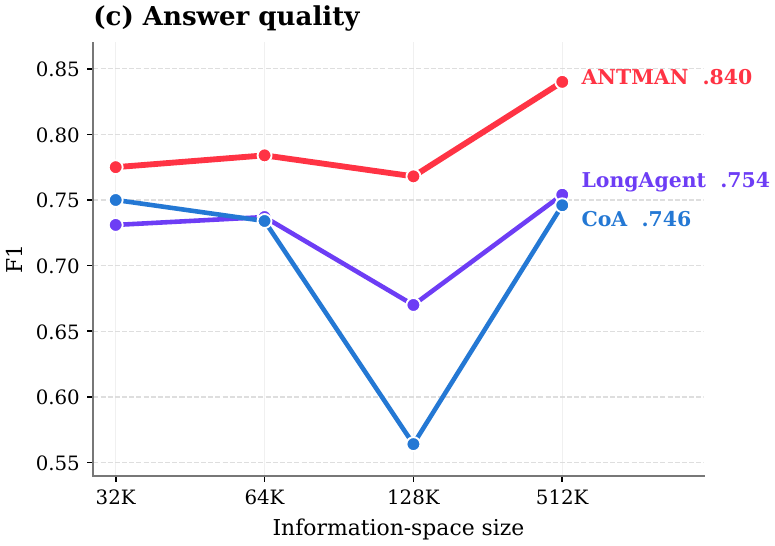}
    \end{minipage}

  \caption{
    \textbf{Selective coordination as information space grows.}
    \textbf{(a)} Active coordination remains nearly constant for ANTMAN
    but grows with space size for LongAgent and CoA.
    \textbf{(b)} This yields slower cost growth.
    \textbf{(c)} Answer quality remains stable across scales.
}
    \label{fig:coordination_scaling}
\end{figure*}

\textbf{Scaling behavior.}
Figure~\ref{fig:coordination_scaling} shows the central scaling advantage
of ANTMAN.
When the information space grows by $16\times$, active coordination
increases by only $1.23\times$, compared with over $15\times$ for
LongAgent and CoA.
This gap carries over to model calls and inference cost, indicating that
ANTMAN avoids expanding coordination simply because more information is
available.

This selectivity does not reduce answer quality.
Importantly, ANTMAN remains stable as the information space grows because
it activates workers according to unresolved needs rather than the size of
the partitioned space. In contrast, LongAgent and CoA expand coordination
as the underlying space is divided into more partitions.
These results support the intended design principle that coordination
should grow with query demand rather than information-space size.
Full scaling results are reported in
Appendix~\ref{app:scaling_details}.

\subsubsection{Controlled Information-Demand Scaling}
\label{sec:demand_scaling}

We complement the space-scaling experiment with the reverse intervention,
holding the searchable space fixed at 512K while increasing required
evidence from 1 to 4 to 16 units. Matched contexts preserve the surrounding
distractors, evidence locations, worker organization, and execution setting.

\textbf{Demand sensitivity.}
Table~\ref{tab:demand_scaling} shows that increasing required evidence from
1 to 16 units raises active coordination from 5.0 to 8.0 workers and model
calls from 35.5 to 86.2 per query. Despite the higher demand, ANTMAN
recovers all required evidence and answers all evaluated queries correctly. Additional construction details and trajectory diagnostics are provided in
Appendix~\ref{app:demand_diagnostics}.
\begin{table}[t]
    \centering
    \scriptsize
    \renewcommand{\arraystretch}{0.86}
    \setlength{\tabcolsep}{3pt}

    \caption{
        \textbf{Coordination under increasing information demand.}
        Searchable space is fixed at 512K while required evidence increases.
    }
    \label{tab:demand_scaling}

    \begin{tabular*}{\linewidth}{
        @{}
        c
        @{\extracolsep{\fill}}
        c
        c
        c
        c
        c
        @{}
    }
        \toprule

        \makecell[c]{\textbf{Required Evidence}\\\textbf{Units}}
        &
        \makecell[c]{\textbf{Active}\\\textbf{Coordination}}
        &
        \makecell[c]{\textbf{Calls}\\\textbf{/ Query}}
        &
        \makecell[c]{\textbf{Cost}\\\textbf{/ Query}}
        &
        \makecell[c]{\textbf{Complete}\\\textbf{Evidence}}
        &
        \makecell[c]{\textbf{Answer}\\\textbf{Acc.}}
        \\

        \midrule

        \rowcolor{OursBand}
        \multicolumn{6}{@{}l@{}}{\hspace{3pt}\textbf{ANTMAN}}
        \\

        \rowcolor{OursBG}
        1
        &
        5.0
        &
        35.5
        &
        \$0.107
        &
        100\%
        &
        100\%
        \\

        \rowcolor{OursBG}
        4
        &
        4.3
        &
        43.6
        &
        \$0.111
        &
        100\%
        &
        100\%
        \\

        \rowcolor{OursBG}
        16
        &
        8.0
        &
        86.2
        &
        \$0.293
        &
        100\%
        &
        100\%
        \\

        \bottomrule
    \end{tabular*}

\end{table}

\begin{rqbox}
\textbf{Answer to RQ1.}
Coordination follows need, not space: ANTMAN keeps its coordination
footprint nearly constant as the searchable space grows, while allocating
substantially more computation when information demand increases at fixed
space. Across both interventions, answer quality remains stable.
\end{rqbox}
\subsubsection{Navigation in Realistic Structured Information Spaces}
\label{sec:structured_navigation}
We evaluate on two repository-level benchmarks and one general
information-seeking benchmark.
RepoProbe-Python~\citep{repoprobe2026} contains 108 questions across
eight large Python repositories, while our frozen 80-question subset of
SWE-QA-Pro~\citep{cai-etal-2026-swe} requires multi-file, agentic
codebase exploration.
GAIA~\citep{ICLR2024_25ae35b5} extends the evaluation to heterogeneous
tool-using information seeking; we use the fixed text-only
GAIA-Text-103 subset.
Within each repository benchmark, methods evaluated in our harness use
the same frozen question set and model setting, while all methods evaluated
on GAIA share the same tool interface
(Appendix~\ref{app:tool_interfaces}).
We additionally include OWL~\citep{hu2025owl} as a general multi-agent
tool-using baseline on GAIA; repository-specific systems are evaluated
on the corresponding code-navigation benchmarks.

\textbf{Structured navigation.}
Table~\ref{tab:structured_navigation} shows that the same need-driven
coordination principle remains effective when evidence must be discovered
through structured navigation.
ANTMAN uses evolving unresolved needs to determine where search should
continue and, in multi-worker substrates, which workers should participate.
Its strong performance across both repository benchmarks shows that this
coordination abstraction transfers beyond the controlled long-context
setting.
The clearest separation from competing methods appears on GAIA-Text-103,
where ANTMAN operates over a distinct web-and-tool substrate.
This provides further evidence that its effectiveness is not tied to
repository-specific navigation machinery. Meanwhile, ANTMAN-H remains close to full ANTMAN, with the main
degradation appearing on more difficult tasks. Most of the GAIA gap comes from Level 3, suggesting that full ANTMAN is most beneficial on harder cases, while ANTMAN-H achieves similar performance on less demanding tasks.

\begin{table*}[t]
    \centering
    \scriptsize
    \renewcommand{\arraystretch}{0.84}
    \setlength{\tabcolsep}{2.0pt}

    \caption{
        \textbf{Performance in realistic structured information spaces.}
        GAIA-Text-103 is broken down by difficulty level.
        Raw repository scores are reported in
        Appendix~\ref{app:structured_raw}.
    }
    \label{tab:structured_navigation}

    \begin{tabular*}{\textwidth}{
        l
        @{\extracolsep{\fill}}
        c c c c c c
    }

        \toprule

        \multirow{2}{*}{\textbf{Method}}
        &
        \multirow{2}{*}{\textbf{RepoProbe}}
        &
        \multirow{2}{*}{\textbf{SWE-QA-Pro}}
        &
        \multicolumn{4}{c}{\textbf{GAIA-Text-103}}
        \\

        \cmidrule(lr){4-7}

        &
        &
        &
        \textbf{L1}
        &
        \textbf{L2}
        &
        \textbf{L3}
        &
        \textbf{Overall}
        \\[-1pt]

        \midrule


        \rowcolor{GroupBG}
        \multicolumn{7}{c}{
            \itshape Retrieval and Iterative Retrieval Baselines
        }
        \\

        Direct
        & 31.08
        & 55.74
        & 28.2
        & 17.3
        & 8.3
        & 20.4
        \\

        Iterative Sparse Retrieval
        & 33.15
        & 60.90
        & 53.8
        & 42.3
        & 33.3
        & 45.6
        \\

        Dense Retrieval
        ~\citep{karpukhin-etal-2020-dense}
        & 31.51
        & 58.30
        & 43.6
        & 26.9
        & 25.0
        & 33.0
        \\

        S2G-RAG
        ~\citep{li-etal-2026-s2g}
        \;\venue{ACL'26}
        & 22.87
        & 60.40
        & 25.6
        & 19.2
        & 0.0
        & 19.4
        \\


        \rowcolor{GroupBG}
        \multicolumn{7}{c}{
            \itshape General Tool-Using Agents
        }
        \\

        ReAct
        ~\citep{yao2023reactsynergizingreasoningacting}
        \;\venue{ICLR'23}
        & 25.86
        & 64.92
        & 46.2
        & 30.8
        & 8.3
        & 34.0
        \\

        OWL
        ~\citep{hu2025owl}
        \;\venue{NeurIPS'25}
        & --
        & --
        & 53.8
        & 38.5
        & 8.3
        & 40.8
        \\


        \rowcolor{GroupBG}
        \multicolumn{7}{c}{
            \itshape Repository-Specific Agents
        }
        \\

        ReAct + RepoGraph
        ~\citep{ICLR2025_4a4a3c19}
        \;\venue{ICLR'25}
        & 25.71
        & 67.38
        & --
        & --
        & --
        & --
        \\

        RepoDistill
        ~\citep{yin-etal-2026-repodistill}
        \;\venue{Findings ACL'26}
        & 28.27
        & 65.10
        & --
        & --
        & --
        & --
        \\


        \rowcolor{GroupBG}
        \multicolumn{7}{c}{
            \itshape Benchmark-Specific Reference
        }
        \\

        SWE-QA-Pro Agent
        ~\citep{cai-etal-2026-swe}
        & --
        & 77.58
        & --
        & --
        & --
        & --
        \\


        \rowcolor{OursBand}
        \multicolumn{7}{c}{
            \itshape Ours
        }
        \\

        \rowcolor{OursBG}
        \textbf{ANTMAN-H$^\dagger$}
        & 34.10
        & 74.60
        & \textbf{69.2}
        & \textbf{55.8}
        & 25.0
        & 57.3
        \\

        \rowcolor{OursBG}
        \textbf{ANTMAN}
        & \textbf{38.21}
        & \textbf{79.78}
        & \textbf{69.2}
        & 53.8
        & \textbf{41.7}
        & \textbf{58.3}
        \\

        \bottomrule

    \end{tabular*}

    \vspace{1pt}

    \begin{minipage}{0.99\textwidth}
        \tiny
        \color{VenueGray}
        Higher is better.
        $^\dagger$ ANTMAN-H uses Qwen3-8B for all non-orchestrator
        components while retaining the same orchestrator as ANTMAN.
        ``--'' denotes settings outside the method-specific evaluation scope.
    \end{minipage}

\end{table*}

\begin{rqbox}
\textbf{Answer to RQ2.}
Need-driven coordination transfers to realistic structured
environments, allowing ANTMAN to selectively navigate repository
and heterogeneous information spaces without requiring a
substrate-specific coordination strategy.
\end{rqbox}

\subsection{Does the Evolving Need State Matter?}
\label{sec:adaptive_need_tracking}

\textbf{RQ3} asks whether ANTMAN benefits from using evolving unresolved
needs as its runtime control state, beyond either a fixed initial plan
or generic adaptive replanning. We compare full ANTMAN with Static ANTMAN,
Graph-free Adaptive, and targeted ablations of Need revision, adaptive
rerouting, and recovery. All variants share the same substrate organization,
models, tools, and execution budget. Appendix~\ref{app:adaptive_ablation}
provides the variant definitions and complete execution statistics.

\begin{table*}[t]
    \centering
    \scriptsize
    \renewcommand{\arraystretch}{0.78}
    \setlength{\tabcolsep}{2.2pt}

    \caption{
        \textbf{Ablation of runtime adaptive coordination.}
        Parentheses indicate relative quality drops from full ANTMAN.
    }
    \label{tab:adaptive_need_results}

    \begin{tabular*}{\textwidth}{
        l
        @{\extracolsep{\fill}}
        c
        c
        c
    }

        \toprule

        \textbf{Variant}
        &
        \makecell[c]{\textbf{512K}\\[-2pt]\textbf{F1}}
        &
        \makecell[c]{\textbf{SWE-QA-Pro}\\[-2pt]\textbf{Score}}
        &
        \makecell[c]{\textbf{GAIA}\\[-2pt]\textbf{Accuracy}}
        \\[-1pt]

        \midrule

        \rowcolor{GroupBG}
        \multicolumn{4}{c}{
            \itshape Adaptive Coordination Ablations
        }
        \\[-1pt]

        Static ANTMAN
        &
        40.32 \AblDrop{52.02}
        &
        68.58 \AblDrop{15.78}
        &
        50.00 \AblDrop{16.67}
        \\

        Graph-free Adaptive
        &
        64.01 \AblDrop{23.82}
        &
        78.88 \AblDrop{3.13}
        &
        50.00 \AblDrop{16.67}
        \\

        w/o Need Revision
        &
        74.44 \AblDrop{11.41}
        &
        75.32 \AblDrop{7.50}
        &
        42.50 \AblDrop{29.17}
        \\

        w/o Adaptive Rerouting
        &
        82.20 \AblDrop{2.18}
        &
        72.13 \AblDrop{11.42}
        &
        60.00 \AblSame{0.00}
        \\

        w/o Recovery
        &
        60.12 \AblDrop{28.45}
        &
        76.52 \AblDrop{6.03}
        &
        52.50 \AblDrop{12.50}
        \\[-1pt]

        \rowcolor{OursBand}
        \multicolumn{4}{c}{
            \itshape Ours
        }
        \\[-1pt]

        \rowcolor{OursBG}
        \textbf{Full ANTMAN}
        &
        \textbf{84.03}
        &
        \textbf{81.43}
        &
        \textbf{60.00}
        \\

        \bottomrule

    \end{tabular*}
\end{table*}

\textbf{Evolving need state and adaptive coordination.}
Table~\ref{tab:adaptive_need_results} shows that full ANTMAN improves
answer quality over both fixed coordination and graph-free adaptive
replanning across all three settings. In controlled scaling, the
Full--Graph-free gap persists despite comparable numbers of activated
workers (Appendix~\ref{app:adaptive_ablation_results}), suggesting that
the benefit of explicit need tracking extends beyond simply engaging
more workers. The targeted ablations further reveal complementary
mechanisms. Recovery is particularly consequential in controlled
scaling, while adaptive rerouting plays an important role in
structured repository navigation. As expected, rerouting has no effect on GAIA, which uses a single territory-backed worker.

\begin{rqbox}
\textbf{Answer to RQ3.}
Explicitly tracking evolving unresolved needs improves
answer quality beyond fixed coordination and graph-free
adaptation, with different runtime mechanisms contributing
across information environments.
\end{rqbox}

\subsection{Position Robustness under Selective Coordination}
\label{sec:position_sensitivity}
\begin{table}[H]
    \centering
    \scriptsize
    \renewcommand{\arraystretch}{0.92}
    \setlength{\tabcolsep}{3.0pt}

    \caption{
        \textbf{Robustness to evidence position as the information space grows.}
        We report F1 when required evidence appears at Early (E),
        Middle (M), or Late (L) positions.
        The final column reports
        $M-\frac{1}{2}(E+L)$ after averaging each position across
        context lengths; negative values indicate a middle-position penalty.
    }
    \label{tab:position_sensitivity}

    \begin{tabular*}{\linewidth}{
        l
        @{\extracolsep{\fill}}
        ccc
        ccc
        ccc
        c
    }

        \toprule

        \multirow{2}{*}{\textbf{Method}}
        &
        \multicolumn{3}{c}{\textbf{32K}}
        &
        \multicolumn{3}{c}{\textbf{64K}}
        &
        \multicolumn{3}{c}{\textbf{128K}}
        &
        \multirow{2}{*}{
            \makecell[c]{
                \textbf{Middle}\\
                \textbf{Gap}
            }
        }
        \\

        \cmidrule(lr){2-4}
        \cmidrule(lr){5-7}
        \cmidrule(lr){8-10}

        &
        E
        &
        \cellcolor{MiddleBG}\textbf{M}
        &
        L
        &
        E
        &
        \cellcolor{MiddleBG}\textbf{M}
        &
        L
        &
        E
        &
        \cellcolor{MiddleBG}\textbf{M}
        &
        L
        &
        \\

        \midrule


        \rowcolor{GroupBG}
        \multicolumn{11}{c}{
            \itshape Direct Long-Context Reference
        }
        \\

        Full Context
        &
        .911
        &
        \cellcolor{MiddleBG}.590
        &
        .813
        &
        .767
        &
        \cellcolor{MiddleBG}.667
        &
        .733
        &
        .733
        &
        \cellcolor{MiddleBG}.613
        &
        .870
        &
        $-.181$
        \\


        \rowcolor{OursBand}
        \multicolumn{11}{c}{
            \itshape Ours
        }
        \\

        \rowcolor{OursBG}
        \textbf{ANTMAN}
        &
        .741
        &
        \cellcolor{OursMiddleBG}.783
        &
        .800
        &
        .747
        &
        \cellcolor{OursMiddleBG}.826
        &
        .780
        &
        .743
        &
        \cellcolor{OursMiddleBG}.748
        &
        .813
        &
        \textbf{+.015}
        \\

        \bottomrule

    \end{tabular*}

    \vspace{1pt}

    \begin{minipage}{0.99\linewidth}
        \tiny
        \color{VenueGray}
        Position results use the same paired questions as the controlled
        scaling experiment.
        Full Context serves as the direct positional-sensitivity reference.
    \end{minipage}

\end{table}

We further examine whether ANTMAN remains robust to where relevant evidence
appears within the information space. Using the same paired Early, Middle,
and Late conditions from Section~\ref{sec:controlled_scaling}, we compare
ANTMAN with direct full-context inference. Table~\ref{tab:position_sensitivity} shows that ANTMAN does not exhibit the
strong positional degradation observed in direct full-context inference.
While Full Context exhibits the classical ``lost in the middle''
pattern~\citep{liu-etal-2024-lost}, ANTMAN remains stable across Early,
Middle, and Late evidence placements. This is consistent with ANTMAN's selective search design, which narrows the
context considered for each unresolved need and thereby reduces sensitivity
to the evidence's position in the global information space. Thus, ANTMAN retains positional
robustness despite activating only a small fraction of the available
workers.

\begin{takeawaybox}
\textbf{Position robustness.}
Selective coordination substantially reduces sensitivity to evidence
position, allowing ANTMAN to remain robust across Early, Middle, and Late
placements.
\end{takeawaybox}

\section{Conclusion}

We introduced ANTMAN, a multi-agent information-seeking
framework that treats evolving unresolved information needs as the runtime
control state for coordination. Rather than organizing computation around
static partitions, ANTMAN maintains and revises a Need Graph that tracks what
remains unresolved and guides information access, worker activation, routing,
and task-local recovery. Across multi-document QA, controlled
information-space scaling, and realistic structured navigation, ANTMAN
maintains strong answer quality while selectively allocating coordination
according to evolving information needs. In particular, active coordination
remains nearly constant as the surrounding information space grows, while
the same need-conditioned mechanism remains effective across different
information-seeking settings and with substantially smaller worker models.
Beyond these results, our findings suggest that coordination need not mirror
the structure of the information space. By making unresolved needs the basis
for runtime control, ANTMAN separates available search capacity from the
computation activated for a query. Looking forward, we aim to extend ANTMAN
to richer information substrates, develop more transferable representations
of information needs, and refine policies for revising and routing those
needs as search progresses.


\bibliography{references}
\bibliographystyle{iclr2027_conference}


\clearpage
\appendix


\section{ANTMAN Execution and Space-Decoupled Coordination}
\label{app:space_decoupling}

This appendix provides the complete execution procedure for ANTMAN and
formalizes why its active coordination is governed by runtime information
needs rather than directly by the size of the underlying information
space.


\subsection{Full Execution Procedure}
\label{app:full_algorithm}

Let $\mathcal{I}$ denote the information space associated with query $q$.
A substrate-specific partitioning procedure $P$ produces a fixed set of
$M$ territories

\begin{equation}
\mathcal{T}
=
P(\mathcal{I})
=
\{\tau_1,\ldots,\tau_M\},
\end{equation}

with corresponding WorkerCards

\begin{equation}
\mathcal{C}
=
\{c_1,\ldots,c_M\},
\qquad
c_i=\mathrm{Card}(\tau_i).
\end{equation}

Each WorkerCard summarizes the scope of its associated territory.
The territory map and WorkerCards remain fixed during execution.
Runtime adaptation occurs only through the Need Graph and the resulting
routing decisions.

Given query $q$, the coordinator initializes a Need Graph
$G_0=(N_0,E_0)$.
Each node $n\in N_t$ represents an information need and maintains
task-local state including resolution status, accumulated evidence,
attempt history, and progress.
Algorithm~\ref{alg:antman} gives the complete execution loop.

\begin{tcolorbox}[
    enhanced,
    colback=white,
    frame hidden,
    borderline north={0.6pt}{0pt}{black!70},
    borderline south={0.6pt}{0pt}{black!70},
    left=4pt,
    right=4pt,
    top=4pt,
    bottom=4pt,
    before skip=6pt,
    after skip=6pt,
    breakable
]

\captionof{algorithm}{
\textsc{ANTMAN} execution procedure.
The substrate map remains fixed, while the Need Graph is revised during
execution to select, route, and recover unresolved information needs.
}
\label{alg:antman}

\vspace{2pt}

\begin{algorithmic}[1]

\Require Query $q$, information space $\mathcal{I}$, execution budget $B$
\Ensure Final answer $y$

\State $\mathcal{T}\gets P(\mathcal{I})$
\Comment{fixed territories}

\State $\mathcal{C}\gets
    \{\mathrm{Card}(\tau_i)\}_{i=1}^{|\mathcal{T}|}$
\Comment{static WorkerCards}

\State $G_0\gets\mathrm{InitializeNeedGraph}(q)$
\State $t\gets 0$

\While{\textbf{not} $\mathrm{Terminate}(G_t,B)$}

    \State $n_t\gets\mathrm{Select}(G_t)$
    \Comment{choose an unresolved need}

    \State $i_t\gets
        \mathrm{Route}(n_t,G_t,\mathcal{C})$
    \Comment{select territory-backed worker}

    \State $r_t\gets
        \mathrm{Exec}(w_{i_t},n_t,\tau_{i_t})$
    \Comment{bounded local information seeking}

    \State $G_{t+1}\gets U(G_t,r_t)$
    \Comment{update evidence, status, and dependencies}

    \If{$\mathrm{Stuck}(n_t,G_{t+1})$}

        \State $a_t\gets
            \mathrm{Recover}(n_t,G_{t+1},\mathcal{C})$

        \If{$a_t=\mathrm{Reframe}$}
            \State $G_{t+1}\gets
                \mathrm{ReframeNeed}(n_t,G_{t+1})$

        \ElsIf{$a_t=\mathrm{Reroute}$}
            \State $\mathrm{MarkForRerouting}(n_t,G_{t+1})$

        \ElsIf{$a_t=\mathrm{Fallback}$}
            \State $G_{t+1}\gets
                \mathrm{Fallback}(n_t,G_{t+1})$
        \EndIf

    \EndIf

    \State $t\gets t+1$

\EndWhile

\State $\mathcal{E}^{*}
    \gets\mathrm{SelectResolvedEvidence}(G_t)$
\State $y\gets\mathrm{Synthesize}(q,\mathcal{E}^{*})$
\State \Return $y$

\end{algorithmic}

\end{tcolorbox}

The key architectural separation in
Algorithm~\ref{alg:antman} is between the static candidate set
$\mathcal{C}$ and the runtime state $G_t$.
Increasing the information-space size may increase
$|\mathcal{C}|=M$, but this does not by itself require more workers
to participate in a query. Workers become active only after an
unresolved need has been selected and dispatched at Line~7.

In our implementation, Line~7 uses a two-stage routing procedure.
A non-LLM retrieval stage ranks WorkerCards using sparse lexical
matching, exact-term matching, and dense similarity, combined with
reciprocal-rank fusion, and provides a bounded set of relevant
candidates for the current routing decision. The orchestrator then
decides whether and where to dispatch the current unresolved need
using the runtime state in $G_t$.

Candidate discovery and runtime coordination play distinct roles.
The retrieval stage limits the WorkerCard information considered
within an individual routing decision, whereas the evolving Need
Graph governs the sequence of routing decisions over the complete
query trajectory, including whether further worker activation is
required and when search can terminate. Our space-decoupling claim
therefore concerns active multi-agent coordination, specifically
whether worker participation is structurally coupled to the number
of available territories, rather than claiming that all retrieval
computation is independent of information-space size.

\subsection{Runtime Coordination Demand}
\label{app:coordination_demand}

To distinguish available search capacity from actual coordination, we
define the unresolved need set at step $t$ as

\begin{equation}
\mathcal{U}_t
=
\{n\in N_t
\mid
s_t(n)\neq \mathrm{resolved}\},
\end{equation}

where $s_t(n)$ denotes the resolution status of need $n$.

Because the Need Graph may be revised during execution, the set of needs
that appear throughout a task can be larger than the initial node set
$N_0$.
We therefore define the task's realized need set as

\begin{equation}
\mathcal{H}(q)
=
\bigcup_{t=0}^{T_q} N_t,
\qquad
H(q)=|\mathcal{H}(q)|,
\label{eq:realized_need_set}
\end{equation}

where $T_q$ is the final coordination step.
A reframed or newly introduced need is counted as a new realized need in
$\mathcal{H}(q)$.

Let $a(n)$ denote the number of routing attempts associated with need
$n$.
This includes its initial routing and any subsequent rerouting caused by
runtime recovery.
Under the execution budget, let

\begin{equation}
a(n)\le \rho
\label{eq:routing_bound}
\end{equation}

for all $n\in\mathcal{H}(q)$, where $\rho$ is the maximum number of
routing attempts permitted for a single need.

The total number of worker-dispatch events is therefore

\begin{equation}
K(q)
=
\sum_{n\in\mathcal{H}(q)} a(n).
\label{eq:coordination_events}
\end{equation}

Finally, let

\begin{equation}
\mathcal{A}(q)
=
\bigcup_{t=0}^{T_q-1}
\{i_t\}
\end{equation}

denote the set of distinct territory-backed workers activated during
execution.

\subsection{Space-Decoupled Coordination}
\label{app:space_decoupling_proof}

We now formalize the sense in which ANTMAN decouples active coordination
from the total number of available territories.

\paragraph{Proposition 1.}
For any query $q$ executed by Algorithm~\ref{alg:antman},

\begin{equation}
|\mathcal{A}(q)|
\le
K(q)
\le
\rho H(q).
\label{eq:coordination_bound_full}
\end{equation}

Consequently,

\begin{equation}
|\mathcal{A}(q)|
\le
\min\{M,\rho H(q)\}.
\label{eq:space_decoupled_bound}
\end{equation}

Thus, the number of workers activated by ANTMAN is bounded by realized
task demand and the per-need routing budget, rather than directly by the
number of available territories $M$.

\paragraph{Proof.}
Every worker activation in Algorithm~\ref{alg:antman} occurs through a
routing decision on Line~7.
Each such decision is associated with one currently selected information
need.

For a realized need $n$, let $a(n)$ be the total number of times it is
routed or rerouted.
Summing these events over all needs encountered during execution gives

\begin{equation}
K(q)
=
\sum_{n\in\mathcal{H}(q)} a(n).
\end{equation}

Since a single routing event can activate at most one
territory-backed worker, the number of \emph{distinct} workers activated
cannot exceed the total number of routing events:

\begin{equation}
|\mathcal{A}(q)|
\le
K(q).
\end{equation}

By Equation~\ref{eq:routing_bound}, each need is routed at most $\rho$
times.
Therefore,

\begin{align}
K(q)
&=
\sum_{n\in\mathcal{H}(q)} a(n)
\\
&\le
\sum_{n\in\mathcal{H}(q)} \rho
\\
&=
\rho |\mathcal{H}(q)|
\\
&=
\rho H(q).
\end{align}

Combining the two inequalities yields

\begin{equation}
|\mathcal{A}(q)|
\le
K(q)
\le
\rho H(q).
\end{equation}

In addition, ANTMAN contains only $M$ territory-backed workers, so
trivially $|\mathcal{A}(q)|\le M$.
Together,

\begin{equation}
|\mathcal{A}(q)|
\le
\min\{M,\rho H(q)\}.
\end{equation}

Importantly, $M$ appears only as the size of the \emph{available}
candidate pool.
It does not appear as a multiplicative factor in the demand-dependent
bound $\rho H(q)$.
Therefore, enlarging the information space and increasing $M$ does not
by itself force additional workers to become active.
\hfill$\square$


\paragraph{Corollary 1.}
Consider a family of increasingly large information spaces
$\{\mathcal{I}_M\}$ for the same class of queries.
If the realized information demand remains bounded such that

\begin{equation}
H(q)\le \bar{H}
\end{equation}

and the per-need routing budget $\rho$ does not depend on $M$, then

\begin{equation}
|\mathcal{A}(q)|
\le
\rho\bar{H}
=
O(1)
\qquad
\text{with respect to } M.
\label{eq:constant_wrt_space}
\end{equation}

Hence ANTMAN does not structurally require active coordination to grow
linearly with information-space size.

\paragraph{Interpretation.}
The result does \emph{not} claim that ANTMAN always activates a constant
number of workers.
A harder query may produce more unresolved needs, more revisions, or
more recovery attempts, increasing $H(q)$ or $K(q)$.
Instead, the proposition states that this growth is induced by
\emph{task demand}.
Simply adding more searchable territories does not automatically add
coordination.


\subsection{Contrast with Partition-Driven Coordination}
\label{app:partition_contrast}

The distinction becomes clearer when compared with a partition-driven
execution rule.
Consider a method that assigns one worker to each of the $M$ partitions
of the information space and activates those workers as part of the
initial decomposition.
Its active worker count satisfies

\begin{equation}
|\mathcal{A}_{\mathrm{partition}}(q)|
=
M,
\end{equation}

or more generally

\begin{equation}
|\mathcal{A}_{\mathrm{partition}}(q)|
=
\Theta(M)
\end{equation}

when a constant fraction of partitions is activated.

ANTMAN instead satisfies

\begin{equation}
|\mathcal{A}_{\mathrm{ANTMAN}}(q)|
\le
\min\{M,\rho H(q)\}.
\end{equation}

The difference is therefore not that ANTMAN ignores the size or
structure of the information space.
A larger space may produce more territories and hence a larger candidate
worker pool.
The difference is that this larger pool is \emph{passive} until the
runtime Need Graph creates demand for it.

Under fixed or comparable information requirements,

\begin{equation}
H(q)
\not\propto
M,
\end{equation}

and therefore ANTMAN does not inherit the linear coordination growth of a
partition-driven strategy.

This is the structural property evaluated in
Section~\ref{sec:controlled_scaling}, where the searchable information
space is increased while the underlying information requirements remain
comparable.

\section{Detailed Controlled Scaling Results}
\label{app:scaling_details}

This section provides additional results for the controlled
information-space scaling experiments in
Section~\ref{sec:controlled_scaling}.
We first summarize endpoint growth from 32K to 512K, then report the
complete per-length results and additional position-sensitivity analysis.


\subsection{Endpoint Scaling}

Table~\ref{tab:scaling_summary} summarizes how coordination, model calls,
cost, and answer quality change from 32K to 512K.

\begin{table}[!htbp]
    \centering
    \scriptsize
    \renewcommand{\arraystretch}{0.92}
    \setlength{\tabcolsep}{3.2pt}

    \caption{
        \textbf{Endpoint scaling from 32K to 512K.}
        Growth factors compare the smallest and largest information
        spaces in the controlled evaluation.
        $\Delta$F1 reports the corresponding absolute change in answer
        quality.
    }
    \label{tab:scaling_summary}

    \begin{tabular*}{\linewidth}{
        @{}
        l
        @{\extracolsep{\fill}}
        c
        c
        c
        c
        @{}
    }

        \toprule

        \textbf{Method}
        &
        \makecell[c]{\textbf{Coord.}\\\textbf{Growth}}
        &
        \makecell[c]{\textbf{Calls}\\\textbf{Growth}}
        &
        \makecell[c]{\textbf{Cost}\\\textbf{Growth}}
        &
        $\Delta$\textbf{F1}
        \\

        \midrule

        \rowcolor{GroupBG}
        \multicolumn{5}{c}{
            \itshape Multi-Agent Baselines
        }
        \\

        LongAgent~\citep{zhao-etal-2024-longagent}
        &
        15.29$\times$
        &
        12.81$\times$
        &
        14.22$\times$
        &
        $+.023$
        \\

        CoA~\citep{zhang2024chainagentslargelanguage}
        &
        15.33$\times$
        &
        11.30$\times$
        &
        16.11$\times$
        &
        $-.004$
        \\

        \rowcolor{OursBand}
        \multicolumn{5}{c}{
            \itshape Ours
        }
        \\

        \rowcolor{OursBG}
        \textbf{ANTMAN}
        &
        \textbf{1.23$\times$}
        &
        \textbf{1.21$\times$}
        &
        \textbf{6.90$\times$}
        &
        \textbf{$+.065$}
        \\

        \bottomrule

    \end{tabular*}
\end{table}


\subsection{Complete Per-Length Results}

Table~\ref{tab:scaling_details} reports the complete results underlying
the scaling curves in Figure~\ref{fig:coordination_scaling}.
Each entry averages 30 evaluations at a given context length
(10 questions $\times$ 3 evidence positions), spanning a
$16\times$ increase in information-space size from 32K to 512K.

\begin{table}[!t]
    \centering
    \scriptsize
    \renewcommand{\arraystretch}{0.92}
    \setlength{\tabcolsep}{3.0pt}

    \caption{
        \textbf{Complete results for controlled information-space scaling.}
        Active coordination denotes the average number of participating
        workers or members per query.
        Calls and cost are averaged per query.
        Lower values are better for coordination, calls, and cost;
        higher values are better for EM and F1.
    }
    \label{tab:scaling_details}

    \begin{tabular*}{\textwidth}{
        @{}
        l
        @{\extracolsep{\fill}}
        c
        c
        c
        c
        c
        c
        @{}
    }

        \toprule

        \textbf{Method}
        &
        \textbf{Length}
        &
        \makecell[c]{\textbf{Active}\\\textbf{Coord.}}
        &
        \makecell[c]{\textbf{Calls}\\\textbf{/ Query}}
        &
        \makecell[c]{\textbf{Cost}\\\textbf{/ Query}}
        &
        \textbf{EM}
        &
        \textbf{F1}
        \\

        \midrule


        \rowcolor{GroupBG}
        \multicolumn{7}{c}{
            \itshape Multi-Agent Baselines
        }
        \\

        \multirow{4}{*}{
            LongAgent~\citep{zhao-etal-2024-longagent}
        }
        &
        32K
        &
        19.00
        &
        22.83
        &
        \$0.0983
        &
        .533
        &
        .731
        \\

        &
        64K
        &
        37.00
        &
        39.50
        &
        \$0.1797
        &
        .533
        &
        .737
        \\

        &
        128K
        &
        73.10
        &
        75.60
        &
        \$0.3549
        &
        .500
        &
        .670
        \\

        &
        512K
        &
        290.50
        &
        292.50
        &
        \$1.3974
        &
        .567
        &
        .754
        \\

        \addlinespace[3pt]

        \multirow{4}{*}{
            CoA~\citep{zhang2024chainagentslargelanguage}
        }
        &
        32K
        &
        5.10
        &
        \textbf{7.10}
        &
        \$0.0809
        &
        .567
        &
        .750
        \\

        &
        64K
        &
        10.00
        &
        \textbf{12.00}
        &
        \$0.1613
        &
        .567
        &
        .734
        \\

        &
        128K
        &
        20.00
        &
        \textbf{22.00}
        &
        \$0.3216
        &
        .400
        &
        .564
        \\

        &
        512K
        &
        78.20
        &
        80.23
        &
        \$1.3027
        &
        .500
        &
        .746
        \\


        \rowcolor{OursBand}
        \multicolumn{7}{c}{
            \itshape Ours
        }
        \\

        \rowcolor{OursBG}
        &
        32K
        &
        \textbf{2.57}
        &
        27.47
        &
        \textbf{\$0.0773}
        &
        \textbf{.600}
        &
        \textbf{.775}
        \\

        \rowcolor{OursBG}
        &
        64K
        &
        \textbf{2.87}
        &
        29.53
        &
        \textbf{\$0.1069}
        &
        \textbf{.667}
        &
        \textbf{.784}
        \\

        \rowcolor{OursBG}
        &
        128K
        &
        \textbf{3.07}
        &
        28.30
        &
        \textbf{\$0.1276}
        &
        \textbf{.600}
        &
        \textbf{.768}
        \\

        \rowcolor{OursBG}
        \multirow{-4}{*}{\textbf{ANTMAN}}
        &
        512K
        &
        \textbf{3.17}
        &
        \textbf{33.10}
        &
        \textbf{\$0.5332}
        &
        \textbf{.667}
        &
        \textbf{.840}
        \\

        \bottomrule

    \end{tabular*}

\end{table}

\section{Raw Scores for Structured Navigation}
\label{app:structured_raw}

Table~\ref{tab:structured_raw} reports the original benchmark-scale
scores underlying the normalized results in
Table~\ref{tab:structured_navigation}.
RepoProbe-Python is reported on its original 10-point scale, while
SWE-QA-Pro is reported on its original 50-point scale.
The main-paper table linearly rescales these values to 0--100.

\begin{table}[!htbp]
    \centering
    \scriptsize
    \renewcommand{\arraystretch}{0.92}
    \setlength{\tabcolsep}{3.2pt}

    \caption{
        \textbf{Raw scores for realistic structured navigation.}
        RepoProbe-Python scores are reported out of 10 and SWE-QA-Pro
        scores out of 50.
    }
    \label{tab:structured_raw}

    \begin{tabular*}{\linewidth}{
        @{}
        l
        @{\extracolsep{\fill}}
        c
        c
        @{}
    }

        \toprule

        \textbf{Method}
        &
        \makecell[c]{
            \textbf{RepoProbe-Python}\\
            \textbf{/ 10}
        }
        &
        \makecell[c]{
            \textbf{SWE-QA-Pro}\\
            \textbf{/ 50}
        }
        \\

        \midrule


        \rowcolor{GroupBG}
        \multicolumn{3}{c}{
            \itshape Retrieval and Iterative Retrieval Baselines
        }
        \\

        Direct
        &
        3.108
        &
        27.87
        \\

        Iterative Sparse Retrieval
        &
        3.315
        &
        30.45
        \\

        Dense Retrieval
        ~\citep{karpukhin-etal-2020-dense}
        &
        3.151
        &
        29.15
        \\

        S2G-RAG
        ~\citep{li-etal-2026-s2g}
        &
        2.287
        &
        30.20
        \\


        \rowcolor{GroupBG}
        \multicolumn{3}{c}{
            \itshape Agentic and Repository-Aware Baselines
        }
        \\

        Matched ReAct
        ~\citep{yao2023reactsynergizingreasoningacting}
        &
        2.586
        &
        32.46
        \\

        ReAct + RepoGraph
        ~\citep{ICLR2025_4a4a3c19}
        &
        2.571
        &
        33.69
        \\

        RepoDistill
        ~\citep{yin-etal-2026-repodistill}
        &
        2.827
        &
        32.55
        \\


        \rowcolor{GroupBG}
        \multicolumn{3}{c}{
            \itshape Benchmark-Specific Reference
        }
        \\

        SWE-QA-Pro Agent
        ~\citep{cai-etal-2026-swe}
        &
        --
        &
        38.79
        \\


        \rowcolor{OursBand}
        \multicolumn{3}{c}{
            \itshape Ours
        }
        \\

        \rowcolor{OursBG}
        \textbf{ANTMAN-H$^\dagger$}
        &
        3.410
        &
        37.30
        \\

        \rowcolor{OursBG}
        \textbf{ANTMAN}
        &
        \textbf{3.821}
        &
        \textbf{39.89}
        \\

        \bottomrule

    \end{tabular*}

    \vspace{1pt}

    \begin{minipage}{0.98\linewidth}
        \tiny
        \color{VenueGray}
        Main-paper scores are obtained by linear normalization:
        RepoProbe-Python $\times 10$ and SWE-QA-Pro $\times 2$.
        $^\dagger$ ANTMAN-H uses Qwen3-8B for all non-orchestrator
        components.
    \end{minipage}

\end{table}

\section{Adaptive-Coordination Ablations}
\label{app:adaptive_ablation}


\subsection{Variant Definitions}
\label{app:adaptive_ablation_definitions}

All variants preserve ANTMAN's territories, WorkerCards, models, tools,
execution budget, and synthesis procedure.
The Need-Graph variants additionally share the same initial decomposition
and initial Need Graph.
Graph-free Adaptive receives the same query, WorkerCards, accumulated
evidence, and interaction history, but does not instantiate or maintain an
explicit persistent representation of unresolved needs.
Instead, the controller may adaptively select workers and revise its next
search action directly from the available interaction history.

Table~\ref{tab:adaptive_ablation_definition} summarizes the state
representation and runtime mechanisms available to each variant.

\begin{table}[!htbp]
    \centering
    \scriptsize
    \renewcommand{\arraystretch}{0.92}
    \setlength{\tabcolsep}{3.0pt}

    \caption{
        \textbf{Definition of state and adaptive-coordination ablations.}
        Checkmarks indicate enabled mechanisms; ``--'' denotes a mechanism
        that is not applicable without an explicit Need Graph.
    }
    \label{tab:adaptive_ablation_definition}

    \begin{tabular*}{\linewidth}{
        @{}
        l
        @{\extracolsep{\fill}}
        c
        c
        c
        c
        @{}
    }

        \toprule

        \textbf{Variant}
        &
        \makecell[c]{\textbf{Explicit}\\\textbf{Need State}}
        &
        \makecell[c]{\textbf{Need}\\\textbf{Revision}}
        &
        \makecell[c]{\textbf{Adaptive}\\\textbf{Rerouting}}
        &
        \textbf{Recovery}
        \\

        \midrule

        Static ANTMAN
        &
        \checkmark
        &
        $\times$
        &
        $\times$
        &
        $\times$
        \\

        Graph-free Adaptive
        &
        $\times$
        &
        --
        &
        \checkmark
        &
        \checkmark
        \\

        w/o Need Revision
        &
        \checkmark
        &
        $\times$
        &
        \checkmark
        &
        \checkmark
        \\

        w/o Adaptive Rerouting
        &
        \checkmark
        &
        \checkmark
        &
        $\times$
        &
        \checkmark
        \\

        w/o Recovery
        &
        \checkmark
        &
        \checkmark
        &
        \checkmark
        &
        $\times$
        \\

        \rowcolor{OursBG}
        \textbf{Full ANTMAN}
        &
        \checkmark
        &
        \checkmark
        &
        \checkmark
        &
        \checkmark
        \\

        \bottomrule

    \end{tabular*}
\end{table}

Static ANTMAN freezes the initial Need Graph and routing plan, disabling
Need revision, adaptive rerouting, and recovery during execution.

Graph-free Adaptive removes the explicit Need Graph while retaining adaptive
control. At each step, the controller observes the query, accumulated
evidence, and prior interaction history and may choose which worker to invoke
next, redirect search based on new evidence, or recover from an unproductive
search direction. It therefore tests whether an explicit evolving
unresolved-need state provides value beyond generic evidence-conditioned
adaptive replanning.

The w/o Need Revision variant retains the initial Need Graph but prevents
changes to its structure or formulation while preserving evidence-dependent
worker selection, rerouting, and recovery.
The w/o Adaptive Rerouting variant allows the Need Graph to evolve but fixes
each need's worker assignment after its initial routing.
The w/o Recovery variant retains Need revision and adaptive rerouting but
disables stuck detection, reframing, and fallback.
Full ANTMAN retains the evolving Need Graph and all three runtime adaptive
mechanisms.

We evaluate all variants under matched execution budgets on the 512K
controlled condition, SWE-QA-Pro, and a fixed 40-question GAIA subset.


\subsection{Complete Results and Analysis}
\label{app:adaptive_ablation_results}

Tables~\ref{tab:ablation_512k}--\ref{tab:ablation_gaia} report answer
quality and execution statistics. Active coordination denotes the
mean number of distinct territory-backed workers activated per query.
It is reported for the multi-worker controlled and repository substrates;
GAIA uses a single territory-backed worker.
All variants receive the same maximum execution budget of 100 LLM calls
per query, but may use that budget differently as a consequence of their
runtime control policy. Realized calls therefore reflect when a variant
continues, revises, or terminates search rather than an independently
allocated resource.


\begin{table}[!htbp]
    \centering
    \tiny
    \renewcommand{\arraystretch}{0.98}
    \setlength{\tabcolsep}{1.2pt}

    \caption{
        \textbf{512K controlled scaling.}
        Answer quality is F1 score; cost and calls are per query.
    }
    \label{tab:ablation_512k}

    \begin{tabular*}{\linewidth}{
        l
        @{\extracolsep{\fill}}
        c
        c
        c
        c
    }

        \toprule

        \textbf{Variant}
        &
        \makecell[c]{\textbf{Answer}\\\textbf{Quality}}
        &
        \makecell[c]{\textbf{Active}\\\textbf{Coordination}}
        &
        \makecell[c]{\textbf{Cost}\\\textbf{(\$/query)}}
        &
        \makecell[c]{\textbf{LLM Calls}\\\textbf{(/query)}}
        \\

        \midrule

        \rowcolor{GroupBG}
        \multicolumn{5}{c}{
            \itshape Adaptive Coordination Ablations
        }
        \\

        Static ANTMAN
        & 40.32 & 1.97 & 0.156 & 28.7
        \\

        Graph-free Adaptive
        & 64.01 & 3.20 & 0.354 & 20.7
        \\

        w/o Need Revision
        & 74.44 & 3.10 & 0.418 & 21.0
        \\

        w/o Adaptive Rerouting
        & 82.20 & 2.00 & 0.451 & 28.2
        \\

        w/o Recovery
        & 60.12 & 2.07 & 0.353 & 25.9
        \\

        \rowcolor{OursBand}
        \multicolumn{5}{c}{
            \itshape Ours
        }
        \\

        \rowcolor{OursBG}
        \textbf{Full ANTMAN}
        & \textbf{84.03}
        & \textbf{3.17}
        & 0.533
        & 33.1
        \\

        \bottomrule

    \end{tabular*}

\end{table}


\begin{table}[!htbp]
    \centering
    \tiny
    \renewcommand{\arraystretch}{0.98}
    \setlength{\tabcolsep}{1.2pt}

    \caption{
    \textbf{SWE-QA-Pro ablation.}
    Results are reported on a 40-question subset sampled with seed 42.
    Answer quality is the direct score; cost and calls are per query.
}
    \label{tab:ablation_sweqa}

    \begin{tabular*}{\linewidth}{
        l
        @{\extracolsep{\fill}}
        c
        c
        c
        c
    }

        \toprule

        \textbf{Variant}
        &
        \makecell[c]{\textbf{Answer}\\\textbf{Quality}}
        &
        \makecell[c]{\textbf{Active}\\\textbf{Coordination}}
        &
        \makecell[c]{\textbf{Cost}\\\textbf{(\$/query)}}
        &
        \makecell[c]{\textbf{LLM Calls}\\\textbf{(/query)}}
        \\

        \midrule

        \rowcolor{GroupBG}
        \multicolumn{5}{c}{
            \itshape Adaptive Coordination Ablations
        }
        \\

        Static ANTMAN
        & 68.58 & 1.35 & 0.273 & 36.5
        \\

        Graph-free Adaptive
        & 78.88 & 2.45 & 0.165 & 22.1
        \\

        w/o Need Revision
        & 75.32 & 3.50 & 0.298 & 32.4
        \\

        w/o Adaptive Rerouting
        & 72.13 & 2.20 & 0.407 & 48.6
        \\

        w/o Recovery
        & 76.52 & 2.15 & 0.522 & 56.5
        \\

        \rowcolor{OursBand}
        \multicolumn{5}{c}{
            \itshape Ours
        }
        \\

        \rowcolor{OursBG}
        \textbf{Full ANTMAN}
        & \textbf{81.43}
        & \textbf{3.00}
        & 0.504
        & 54.3
        \\

        \bottomrule

    \end{tabular*}

\end{table}


\begin{table}[!htbp]
    \centering
    \tiny
    \renewcommand{\arraystretch}{0.98}
    \setlength{\tabcolsep}{1.2pt}

    \caption{
        \textbf{GAIA ablation (40 questions).}
        Answer quality is accuracy; cost and calls are per query.
    }
    \label{tab:ablation_gaia}

    \begin{tabular*}{\linewidth}{
        l
        @{\extracolsep{\fill}}
        c
        c
        c
    }

        \toprule

        \textbf{Variant}
        &
        \makecell[c]{\textbf{Answer}\\\textbf{Quality}}
        &
        \makecell[c]{\textbf{Cost}\\\textbf{(\$/query)}}
        &
        \makecell[c]{\textbf{LLM Calls}\\\textbf{(/query)}}
        \\

        \midrule

        \rowcolor{GroupBG}
        \multicolumn{4}{c}{
            \itshape Adaptive Coordination Ablations
        }
        \\

        Static ANTMAN
        & 50.00 & 0.196 & 42.8
        \\

        Graph-free Adaptive
        & 50.00 & 0.179 & 54.0
        \\

        w/o Need Revision
        & 42.50 & 0.115 & 28.7
        \\

        w/o Adaptive Rerouting
        & 60.00 & 0.221 & 47.4
        \\

        w/o Recovery
        & 52.50 & 0.325 & 69.5
        \\

        \rowcolor{OursBand}
        \multicolumn{4}{c}{
            \itshape Ours
        }
        \\

        \rowcolor{OursBG}
        \textbf{Full ANTMAN}
        & \textbf{60.00}
        & 0.206
        & 45.4
        \\

        \bottomrule

    \end{tabular*}

\end{table}


\paragraph{Explicit need state and worker activation.}
In controlled scaling, full ANTMAN achieves substantially higher
answer quality than Graph-free Adaptive while activating a comparable
number of distinct workers. This suggests that the benefit of an
explicit evolving Need Graph extends beyond simply engaging a larger
worker set, highlighting the role of persistent unresolved-need
tracking in directing subsequent search.

\paragraph{Need revision and adaptive rerouting.}
On SWE-QA-Pro, disabling Need revision lowers answer quality despite
activating more distinct workers than full ANTMAN, indicating that
broader worker activation alone does not ensure better evidence
acquisition. Disabling adaptive rerouting also reduces answer quality,
highlighting the value of revisiting worker assignments during
repository navigation.

\paragraph{Recovery and execution cost.}
Removing recovery substantially reduces answer quality in controlled
scaling. On GAIA, the same ablation lowers accuracy while increasing
both model calls and inference cost. Thus, recovery contributes to
answer quality and, in this setting, is also associated with lower
realized execution cost.

\section{Primary Runtime Behaviors}
\label{app:runtime_prompts}

This appendix summarizes the primary LLM-guided behaviors used during
ANTMAN's runtime coordination loop. We focus on Need Graph construction
and revision, need-conditioned worker execution, evidence selection, and
need resolution.


\subsection{Orchestrator}

\noindent
\colorbox{OursBand}{%
\parbox{\dimexpr\linewidth-2\fboxsep\relax}{%
\small
\textbf{Need planning and selection.}
Given the current query, Need Graph, accumulated evidence, and execution
history, identify the unresolved information requirements that should be
pursued next. Revise or decompose needs when necessary, respect dependencies
between needs, and select suitable workers for needs that are ready for
execution.
}}

\vspace{4pt}

\noindent
\colorbox{OursBG}{%
\parbox{\dimexpr\linewidth-2\fboxsep\relax}{%
\small
\textbf{Need Graph consolidation.}
Given the current Need Graph and a proposed new or revised need, determine
how it should be incorporated into the graph. The proposal may be added as a
new requirement, attached to an existing dependency, merged with an
overlapping need, subsumed by an existing need, or discarded if redundant.
}}

\vspace{4pt}


\subsection{Worker Execution}

\noindent
\colorbox{OursBand}{%
\parbox{\dimexpr\linewidth-2\fboxsep\relax}{%
\small
\textbf{Need-conditioned action planning.}
Given a selected unresolved need, the worker's assigned territory, and the
available substrate-specific tools, choose and order a bounded sequence of
actions for locating evidence relevant to the current need. The worker
operates locally and returns evidence, remaining uncertainty, and progress
to the coordinator.
}}

\vspace{4pt}


\subsection{Evidence and Need-State Update}

\noindent
\colorbox{OursBand}{%
\parbox{\dimexpr\linewidth-2\fboxsep\relax}{%
\small
\textbf{Evidence selection.}
Given the evidence accumulated during execution, retain the evidence most
relevant to resolving the query and discard redundant or unsupported
material. The selected evidence is passed to subsequent coordination and
final answer synthesis.
}}

\vspace{4pt}

\noindent
\colorbox{OursBG}{%
\parbox{\dimexpr\linewidth-2\fboxsep\relax}{%
\small
\textbf{Need-resolution assessment.}
Given the current need and the evidence collected for it, determine whether
the need is \texttt{resolved}, \texttt{partial}, or \texttt{unresolved}.
A resolved need is closed; an unresolved need remains active; and a partial
need is replaced or refined into a more specific unresolved requirement for
subsequent coordination.
}}


\section{Additional Experimental Details}
\label{app:additional_experimental_details}


\subsection{Substrate-Specific Tool Interfaces}
\label{app:tool_interfaces}

ANTMAN retains the same need-conditioned coordination abstraction across
environments while adapting the information-access layer to each
substrate. The controlled long-context setting uses lexical and semantic
retrieval together with bounded document reading, whereas repository
navigation additionally exposes structural code relations. GAIA uses a
separate web-oriented interface for external information seeking and
lightweight computation. Table~\ref{tab:tool_interfaces} summarizes these
capabilities at the interface level.

\begin{table}[!t]
    \centering
    \scriptsize
    \renewcommand{\arraystretch}{0.92}
    \setlength{\tabcolsep}{3.0pt}

    \caption{
        Information-access capabilities used across evaluation substrates.
    }
    \label{tab:tool_interfaces}

    \begin{tabular*}{\columnwidth}{
        @{}
        p{0.30\columnwidth}
        @{\extracolsep{\fill}}
        p{0.64\columnwidth}
        @{}
    }
        \toprule

        Setting
        &
        Information access
        \\

        \midrule

        Controlled long context
        &
        Lexical and semantic retrieval with bounded reading of selected
        document regions.
        \\

        Repository navigation
        &
        Retrieval and bounded reading augmented with structural navigation
        over symbols, imports, references, inheritance, and call relations.
        \\

        GAIA-Text-103
        &
        Web search, webpage text extraction, and isolated Python execution
        for tool-assisted information seeking.
        \\

        \bottomrule
    \end{tabular*}

\end{table}

The controlled long-context and repository evaluations share the same
local information-access implementation, with the repository setting
additionally supporting structural code navigation. This allows workers
to follow relations among program entities as well as retrieve and inspect
relevant content. These substrate-specific capabilities affect how evidence
is accessed within a territory, while the Need Graph, worker organization,
and need-conditioned coordination procedure remain unchanged.

For GAIA-Text-103, all evaluated methods receive the same tool interface.
The shared interface supports
text-oriented web search and webpage access together with isolated Python
execution. Because GAIA-Text-103 contains no task attachments,
attachment-oriented capabilities are not used in this evaluation. Web
search is backed by Tavily.


\subsection{Controlled Information-Demand Scaling Details}
\label{app:demand_diagnostics}

\paragraph{Task construction.}
We construct 10 matched 512K base contexts, each containing 16 planted
evidence records distributed across a fixed pool of eight territories.
Multiple records may occur in the same territory. For each base context,
the document collection, distractors, evidence locations, territory
assignment, and WorkerCards are held fixed across all demand conditions.
Only the query and corresponding gold answer change.

Each evidence record contains an independently generated value
$e_i \in \{0,\ldots,7\}$. For required-evidence level
$k\in\{1,4,16\}$, the answer is determined by
\[
    y_k =
    \left(
        \sum_{e_i \in D_k} e_i
    \right)
    \bmod 8,
\]
where $D_1 \subset D_4 \subset D_{16}$ are nested required-evidence sets.
The eight outcomes are mapped to neutral answer labels. Because each
required value is independently generated, omitting any required evidence
unit leaves the answer ambiguous over all eight classes.

We evaluate 30 matched conditions in total
(10 base contexts $\times$ 3 demand levels). All conditions use the same
model, tools, routing mechanism, worker organization, and execution budget,
with a maximum of 10 coordination rounds. Answer quality is measured by
exact-match accuracy, while \emph{Complete Evidence} records whether all
required evidence units are recovered. Evidence recovery is determined from
the extracted evidence content rather than from merely accessing the
corresponding document.

\paragraph{Low-demand exploration overhead.}
Although active coordination decreases slightly from 5.0 workers at one
required unit to 4.3 at four units, trajectory analysis shows that this
effect is driven by exploratory overhead rather than failed evidence
acquisition. Table~\ref{tab:demand_diagnostics} decomposes trajectory-level
worker activity according to whether it contributes to required evidence
acquisition.

\begin{table}[!htbp]
    \centering
    \scriptsize
    \renewcommand{\arraystretch}{0.92}
    \setlength{\tabcolsep}{3.0pt}

    \caption{
        \textbf{Trajectory diagnostics under increasing information demand.}
        Worker activity is decomposed according to whether it contributes to
        required evidence acquisition.
    }
    \label{tab:demand_diagnostics}

    \begin{tabular*}{\linewidth}{
        @{}
        l
        @{\extracolsep{\fill}}
        c
        c
        c
        @{}
    }
        \toprule

        \textbf{Required Evidence}
        &
        \makecell[c]{\textbf{Worker}\\\textbf{Activity}}
        &
        \makecell[c]{\textbf{Relevant}\\\textbf{Activity}}
        &
        \makecell[c]{\textbf{Irrelevant}\\\textbf{Activity}}
        \\

        \midrule

        \rowcolor{OursBand}
        \multicolumn{4}{l}{\textbf{ANTMAN}}
        \\

        \rowcolor{OursBG}
        1 unit
        &
        5.0
        &
        0.9
        &
        4.1
        \\

        \rowcolor{OursBG}
        4 units
        &
        4.3
        &
        3.2
        &
        1.1
        \\

        \rowcolor{OursBG}
        16 units
        &
        8.2
        &
        8.0
        &
        0.2
        \\

        \bottomrule
    \end{tabular*}

\end{table}

At one required evidence unit, 4.1 of 5.0 worker activities are unrelated
to the required evidence, accounting for 82\% of the observed activity.
At four units, 3.2 of 4.3 activities contribute to required evidence
acquisition, while irrelevant activity falls to 1.1. At sixteen units,
nearly all observed worker activity contributes to required evidence
acquisition. Thus, the slight low-demand inversion in active coordination
is explained by proportionally greater exploration overhead at small
information demand, while worker activity becomes increasingly aligned
with required evidence as demand grows. Complete evidence recovery and
answer accuracy remain at 100\% across all three demand levels.

\subsection{RepoProbe-Python Execution Diagnostics}
\label{app:repoprobe_execution}

We further examine whether the RepoProbe-Python results can be explained
by differences in execution budget or by the amount of repository
interaction. All methods use the same frozen question set and model
setting. Agentic methods are configured with the same maximum budget of
100 LLM calls per query. Table~\ref{tab:repoprobe_execution} reports
realized LLM calls, tool calls, generation cost, and answer quality.

\begin{table}[!htbp]
    \centering
    \scriptsize
    \renewcommand{\arraystretch}{0.92}
    \setlength{\tabcolsep}{2.6pt}

    \caption{
        \textbf{Execution diagnostics on RepoProbe-Python.}
        LLM calls, tool calls, and generation cost are averaged per query.
        RepoProbe-Python scores are reported on the same 0--100 scale as
        Table~\ref{tab:structured_navigation}.
    }
    \label{tab:repoprobe_execution}

    \begin{tabular*}{\linewidth}{
        @{}
        l
        @{\extracolsep{\fill}}
        c
        c
        c
        c
        @{}
    }

        \toprule

        \textbf{Method}
        &
        \makecell[c]{\textbf{LLM}\\\textbf{Calls}}
        &
        \makecell[c]{\textbf{Tool}\\\textbf{Calls}}
        &
        \makecell[c]{\textbf{Cost}\\\textbf{/ Query}}
        &
        \makecell[c]{\textbf{RepoProbe}\\\textbf{Score}}
        \\

        \midrule


        \rowcolor{GroupBG}
        \multicolumn{5}{c}{
            \itshape Retrieval and Iterative Retrieval Baselines
        }
        \\

        Direct
        &
        1.00
        &
        0.00
        &
        \$0.0067
        &
        31.08
        \\

        Iterative Sparse Retrieval
        &
        3.19
        &
        2.16
        &
        \$0.0217
        &
        33.15
        \\

        Dense Retrieval
        ~\citep{karpukhin-etal-2020-dense}
        &
        1.00
        &
        1.00
        &
        \$0.0124
        &
        31.51
        \\

        S2G-RAG
        ~\citep{li-etal-2026-s2g}
        &
        37.12
        &
        24.00
        &
        \$0.1788
        &
        22.87
        \\


        \rowcolor{GroupBG}
        \multicolumn{5}{c}{
            \itshape Agentic and Repository-Aware Baselines
        }
        \\

        Matched ReAct
        ~\citep{yao2023reactsynergizingreasoningacting}
        &
        18.12
        &
        16.98
        &
        \$0.1136
        &
        25.86
        \\

        ReAct + RepoGraph
        ~\citep{ICLR2025_4a4a3c19}
        &
        26.62
        &
        25.09
        &
        \$0.1922
        &
        25.71
        \\

        RepoDistill
        ~\citep{yin-etal-2026-repodistill}
        &
        4.18
        &
        35.22
        &
        \$0.0801
        &
        28.27
        \\


        \rowcolor{OursBand}
        \multicolumn{5}{c}{
            \itshape Ours
        }
        \\

        \rowcolor{OursBG}
        \textbf{ANTMAN}
        &
        38.13
        &
        26.81
        &
        \$0.2987
        &
        \textbf{38.21}
        \\

        \bottomrule

    \end{tabular*}

\end{table}

\paragraph{Execution diagnostics.}
All agentic RepoProbe-Python methods use the same maximum budget of
100 LLM calls per query, while average realized usage remains below
40 calls for every method (Table~\ref{tab:repoprobe_execution}).
Realized interaction count also does not exhibit a simple monotonic
relationship with performance. In particular, S2G-RAG and ANTMAN use
similar numbers of LLM calls (37.12 vs.\ 38.13) and tool calls
(24.00 vs.\ 26.81), yet obtain substantially different RepoProbe-Python
scores (22.87 vs.\ 38.21). Thus, the observed performance differences
cannot be explained by unequal externally imposed call limits or by
interaction count alone.

\paragraph{Iterative sparse retrieval.}
The sparse-retrieval baseline is a lightweight iterative lexical
retrieval method rather than a one-shot BM25 reader. It allows up to
three retrieval rounds, with the LLM determining whether the accumulated
evidence is sufficient and reformulating the lexical query when further
retrieval is needed. Its average of 2.16 tool calls and 3.19 LLM calls
per query reflects early termination on a subset of questions.

\end{document}